\documentclass{article}

\usepackage{PRIMEarxiv}
\usepackage[utf8]{inputenc} 
\usepackage[T1]{fontenc}    
\usepackage{hyperref}       
\usepackage{url}            
\usepackage{booktabs}       
\usepackage{amsfonts}       
\usepackage{nicefrac}       
\usepackage{microtype}      
\usepackage{lipsum}
\usepackage{fancyhdr}       
\usepackage{graphicx}       
\graphicspath{{media/}}     

\usepackage{natbib}
\usepackage{amsmath}
\usepackage{float}
\usepackage{booktabs} 
\usepackage{appendix}
\usepackage{multirow}

\title{DUNE: A Machine Learning Deep UNet++ based Ensemble Approach to Monthly, Seasonal and Annual Climate Forecasting}

\author{
  Pratik Shukla, Milton Halem \\
  Department of Computer Science and Electrical Engineering \\
  University of Maryland Baltimore County \\
  Baltimore, MD, USA\\
  \texttt{\{pratiks2, halem\}@umbc.edu} \\
}

\begin{document}
\maketitle

\begin{abstract}
Capitalizing on the recent availability of ERA5 monthly averaged long-term data records of mean atmospheric and climate fields based on high-resolution reanalysis, deep-learning architectures offer an alternative to physics-based daily numerical weather predictions for subseasonal to seasonal (S2S) and annual means. A novel Deep UNet++-based Ensemble (DUNE) neural architecture is introduced, employing multi-encoder-decoder structures with residual blocks. When initialized from a prior month or year, this architecture produced the first AI-based global monthly, seasonal, or annual mean forecast of 2-meter temperatures (T2m) and sea surface temperatures (SST). ERA5 monthly mean data is used as input for T2m over land, SST over oceans, and solar radiation at the top of the atmosphere for each month of 40 years to train the model. Validation forecasts are performed for an additional two years, followed by five years of forecast evaluations to account for natural annual variability. AI-trained inference forecast weights generate forecasts in seconds, enabling ensemble seasonal forecasts. Root Mean Squared Error (RMSE), Anomaly Correlation Coefficient (ACC), and Heidke Skill Score (HSS) statistics are presented globally and over specific regions. These forecasts outperform persistence, climatology, and multiple linear regression for all domains. DUNE forecasts demonstrate comparable statistical accuracy to NOAA's operational monthly and seasonal probabilistic outlook forecasts over the US but at significantly higher resolutions. RMSE and ACC error statistics for other recent AI-based daily forecasts also show superior performance for DUNE-based forecasts. The DUNE model's application to an ensemble data assimilation cycle shows comparable forecast accuracy with a single high-resolution model, potentially eliminating the need for retraining on extrapolated datasets.
\end{abstract}



\section{Introduction}
\label{sec:Introduction}
Traditional numerical weather prediction (NWP) models employed by operational weather forecasting centers for subseasonal to seasonal (S2S) forecasts are based on ensembles of physics-based daily weather forecasting models. Operational models aim to overcome extended-range predictability limitations by using the mean of ensembles of forecast models with either different physical parameterizations or perturbations of the initial forecast conditions. All such ensemble members contain variations in their numerical approximations of fluid flows and physical parameterizations of sub-scale meteorological processes. Nevertheless, NWP has been shown to have limited forecast skill beyond two weeks, regardless of model parameterization and resolution of initial observations \citep{realbutterfly, feasibilityexp, lorenz1963predictability, halem1970analysis, lorenz1982atmospheric, zhang2019predictability}. Machine learning has now demonstrated similar 2-4 week forecast skills \citep{shen2023lorenz, kirtmanstatement, weyn2021sub, bi2023accurate, nguyen2023climax} based on training with hourly to daily reanalysis data, yet it suffers from the same limitations in extended predictability beyond two weeks \citep{realbutterfly}. Since boundary forcings govern long-term seasonal and annual climates based on mean properties of the atmosphere, the problem of S2S prediction is addressed using a deep learning self-generating AI ensemble forecasting model, Deep UNet++ Ensemble (DUNE), trained on very long records of reanalysis observations of monthly mean ERA5 data. This study focuses on Subseasonal-to-Seasonal and Annual (S2SA) machine learning model forecasting.

Reliable predictive models of Subseasonal to Seasonal (S2S) temperatures at 2 meters above the surface (T2m) are helpful and often urgently needed for many environmental, economic, and social applications. Temperatures, precipitation, and potential evaporation are indices of drought and significant components influencing wildfire behavior. To enhance mega wildfire likelihood forecasts, one must predict surface drought indices such as annual surface temperature heat anomalies, potential evapotranspiration influenced by T2m, and vegetation dryness due to extended dry precipitation anomalies. Such accurate predictions are vital for anticipating wildfire likelihood and severity. Predicting seasonal and annual surface temperature variations is equally critical. Higher summer temperatures in Boreal regions signal broader climate change trends. Understanding the complex relationship between drought conditions and wildfires is crucial for refining predictive models. Droughts create favorable ignition conditions and contribute to vegetation drying, enhancing wildfire intensity and extent. Accurate drought predictions will enable actionable strategies for proactive wildfire management. Thus, as a vital first step in implementing sustainable strategies to address the challenges of global warming's impact on wildfire risk, we focus in this paper mainly on developing a comprehensive approach to predicting summer and annual T2m. Such an AI model is applicable to the operational requirement of subseasonal to seasonal prediction.

In the \textbf{Related Work} section \ref{sec: Related Work}, we review relevant ML papers, including numerical weather prediction (NWP) forecasts and various short-term and long-term AI-based approaches. The \textbf{Data Description} section \ref{sec: Data Description} details the datasets utilized for training our model. An outline of the methods employed to assess our model's performance are presented in the \textbf{Evaluation} section \ref{sec: Evaluation}. The \textbf{Methodology} section \ref{sec: Methodology} provides an in-depth description of the DUNE AI model architecture and its training procedures. In the \textbf{Results} section \ref{sec:DUNE Forecast Results}, forecasts for monthly, seasonal, and yearly means, as well as moving window forecasts are presented. The \textbf{Discussions} section \ref{sec: Discussions} offers insights and reflections on the experiments conducted. Finally, the \textbf{Conclusions} section \ref{sec: Conclusion} provides closing remarks and summaries. The Appendix provides supplementary graphs, tables, and plots. 

\section{Related Work}
\label{sec: Related Work}
Weather forecasts play a crucial role in our modern world, impacting various sectors of the economy and society. Accurate temperature predictions are critical for agriculture, water management, energy production, extreme heat, drought and wildfire risk.

Neural networks are emerging as complementary models and even as an alternative to operational numerical weather prediction (NWP) models. The recent availability of reanalysis observational data along with computational capabilities has facilitated the development of ensemble probabilistic forecasting methodologies for NWP models \citep{li2022convolutional, rasp2018neural,gronquist2021deep}. These methodologies leverage deep neural networks to generate and improve the representation of forecast uncertainty, which is critical for predictability under uncertainty.

\citet{larraondo2019data} explored various convolutional neural network (CNN) architectures for predicting precipitation fields from geopotential height fields using reanalysis data. Their findings showed the superiority of the U-Net architecture in generating more accurate precipitation forecasts, thus suggesting this CNN algorithm for our weather prediction endeavors.

Additionally, \citep{weyn2019can, weyn2020improving, weyn2021sub} utilized the Resnet CNN to forecast atmospheric variables over different time scales, surpassing climatological benchmarks for weather prediction. Similar experiments involving CNNs by \citet{taylor2022deep} highlighted the role of land surface temperatures (T2m) and sea surface temperatures (SST) in advancing temperature forecasting. Their study primarily focused on analyzing sea surface temperature anomalies using monthly mean data to predict ocean temperatures. It is well-established that weather forecasts, particularly for T2m, tend to exhibit lower accuracy beyond two weeks. Inspired by \citet{taylor2022deep}, we departed from the conventional approach of using daily data and adopted monthly mean data for our experiments. Moreover, we incorporated a mixed field of T2m and SST, aligning with the approach presented by \citep{taylor2022deep}.

In recent years, several innovative approaches have emerged for weather forecasting. However, most of the papers focused on short-term forecasts ranging from a few hours to a few weeks. In their paper, "FourCastNet," \citet{pathak2022fourcastnet} implemented the Adaptive Fourier Neural Operator (FNO) framework by \citet{guibas2021adaptive} for weather forecasting, efficiently capturing complex spatiotemporal patterns. \citet{lam2022graphcast} introduced "GraphCast," a high-resolution graph neural network model designed for 10-day global weather forecasting, outperforming conventional models. In "ClimaX," \citep{nguyen2023climax} utilized the transformer architecture with innovative encoding and aggregation blocks for weather forecasting, enabling accurate predictions across different geographic regions and time scales. 

Numerous researchers have demonstrated proficiency in utilizing the UNet architecture for weather forecasting. \citet{unal2023climate} utilized the UNet++ architecture proposed by \citep{zhou2018unet++} and emphasized the significance of using anomalies over raw data to enhance forecasts specifically for long-term weather forecasting for multi-annual forecasts. The UNet++ architecture developed by \citep{zhou2018unet++}, was shown to significantly advance in medical applications over the UNet model. UNet++ incorporates skip connections at multiple scales, which helps in better integrating feature maps and allows information flow between different layers. This hierarchical connectivity is essential as it enables the model to capture both local and global context, improving its ability to understand complex patterns in the data. Additionally, UNet++ introduces a nested architecture where each encoder block is paired with a corresponding decoder block, further enhancing feature extraction and semantic segmentation performance.

Thus, the UNet++ architecture was tailored to suit the DUNE inference needs. Finally, a new ensemble mean forecast approach was employed by averaging four intermediate forecast outcomes generated by the DUNE AI model. This ensemble technique proved effective, especially during fall and winter, resulting in better forecast accuracy where there was considerable variability in the observations. See the Methodology section \ref{sec: Methodology}.

\textbf{Key Contributions}
\begin{itemize}
    \item Assembled a novel deep learning architecture called the Deep UNet++-based Ensemble (DUNE), specifically designed to incorporate intermediate ensemble inferences (i.e., forecasts) to account for the natural variability in S2S predictions for a range of global climate parameters, including T2m, SST, U10, V10, and more.
    \item Presented AI-based forecasts for monthly, seasonal, and annual means, validated against monthly ground truth (Reanalysis) data for the past five years (2019-2023). These forecasts encompass global regions, land areas, Oceans, the US, Australia, and the global Boreal regions. Additionally, we have exhibited the feasibility of using DUNE in an operational setting by producing forecasts for 2024 and 2025.
    \item Produced the first AI-based 12-month forecast for 2024, comparable statistically to NOAA's operational 12-month probabilistic forecasts, 1000 times faster. 
    \item Trained the DUNE AI model on T2m over land and SST over oceans, enabling simultaneous forecasting of both SST and T2m in a single forecast.
    \item Employed the DUNE AI model, explicitly trained on anomalies, to predict anomalies rather than actual temperatures.
    \item Delivered the first Boreal summer anomaly forecast for the mega fire years 2021 and 2023.
    \item Demonstrated the use of intrinsic ensembles within the DUNE AI model, which improves forecast accuracy during the Fall and Winter months when temperature variability is high.
    \item Illustrated the global temperature increase over the past 5 years by presenting month-by-month forecasts relative to 20-year temperature intervals.
    \item Experimented with integrating wavelet transforms into the downsampling component of our DUNE AI architecture. Given their promising results, we plan to implement trainable wavelet transforms in the DUNE AI model's downsampling and upsampling phases. Due to increased training time, wavelet transforms were not used in the current DUNE AI architecture.
\end{itemize}

\section{Data Description}
\label{sec: Data Description}
The historical atmospheric data for training the AI prediction model is sourced from the European Center for Medium-Range Weather Forecasts (ECMWF) monthly mean ReAnalysis version 5, which we obtained from the Copernicus Climate Change Service (C3S) \citep{hersbach2020era5}. This ERA5 reanalysis dataset offers monthly estimates of various atmospheric, land, and oceanic variables worldwide, a spatial resolution of 0.25\textdegree, roughly equivalent to 30 km. It also includes a 10-member ensemble at roughly 60km. ECMWF developed ERA5 with a global 4-D variational data assimilation cycle using a short-term physical integration as a background and the assimilation of all the available ground and space-borne observations and their error statistics, creating a globally comprehensive and consistent dataset based on current physical and chemistry parameterizations. The dataset covers the Earth and has dimensions of 721 x 1440 for latitude and longitude, respectively. The ensemble data set is 360 x720.  The monthly mean data is available from January 1940 onwards with a temporal resolution of 1 month.

A combination of sea surface temperature (ERA5 label ``sst'') and 2m temperature (ERA5 label ``t2m'') variables from the ERA5 data set is used as input for training. The land-sea mask (ERA5 label ``lsm'') parameter is provided by the ERA5 data, which was used to combine the SST and T2m data sets. 

The following five constants plus solar radiation fields are used as input to train the AI model as suggested by \citet{weyn2020improving}. The constants are lsm, orography (ERA5 label ``z''/9.80665 ), Soil Type (ERA5 label``slt''), High Vegetation Cover (ERA5 label ``cvh'') and Low Vegetation Cover (ERA5 label ``cvl''). Lastly, we used the Top of the Atmosphere Incident Solar Radiation (ERA5 label ``tisr''), which shows the incoming solar radiation from the sun at the top of the atmosphere. This is the only value that 
varies monthly, although it remains constant over the years.  

The complete ERA5 monthly data set was downloaded from January 1940 to June 2024. To calculate the climatology for each month (January to December), the temperature data used spans the 30-year period, from January 1950 to December 1979. An anomaly temperature data set is formed by subtracting the climatology of the respective month from the respective monthly value for each month. Instead of predicting the actual temperature values, the model is trained to predict the temperature anomalies. The actual temperature data are recovered by adding the climatological averages for the respective month.

In the following experiments, data from January 1980 to December 2016 (444 months) are used as input for training, January 2017 to December 2018 (24 months) is used for validation, and January 2019 to December 2023 (60 months) is for conducting the experiments. All the inputs are normalized. A standard normalization technique was applied as shown in equation \eqref{eq: normalization}.The following scaling formula was used for the six constant fields to normalize each variable separately. Since the model predicts the scaled temperature, an inverse scaling was applied to the model's output. The unscaled output was used for model validation.

\begin{equation}
z_i = \frac{{x_i - x_{\text{min}}}}{{x_{\text{max}} - x_{\text{min}}}} \label{eq: normalization}
\end{equation}

In equation \eqref{eq: normalization},  \( z_i \) is the normalized parameter value, \( x_i \) is the raw or the input parameter value, \(x_\text{min}\) is the minimum parameter value, and \(x_\text{max}\) is the maximum parameter value.

\section{Evaluation}
\label{sec: Evaluation}

The loss function is adopted from \cite{rasp2020weatherbench}. The root mean squared error (RMSE) is employed as our primary evaluation metric because it penalizes large differences in the predictions. 

\begin{equation}
\text{RMSE} = \frac{1}{N_{\text{forecasts}}}\sum_{i=1}^{N_{\text{forecasts}}} \sqrt{ \frac{1}{N_{\text{lat}} N_{\text{lon}}} \sum_{j=1}^{N_{\text{lat}}} \sum_{k=1}^{N_{\text{lon}}} L(j) \left(\text{forecast}_{i,j,k} - \text{truth}_{i,j,k}\right)^2 \label{eq: RMSE}}
\end{equation}

In equation \eqref{eq: RMSE}, \(forecast\) is the model forecast, \(truth\) is the ERA5 ground truth, \(N_\text{lat}\) represents the number of latitude values, \(N_\text{lon}\) represents the number of longitude values, \(N_\text{forecasts}\) represents the total number of forecasts made, and \(L(j)\) represents the mean latitude weighting factor.

A mean-latitude-weighting factor is used because the area covered by a 0.25\textdegree x 0.25\textdegree grid box decreases as we move from the tropics to the polar regions. It's important to note that the grid box area remains constant for different longitude values at each latitude. The mean-latitude-weighting factor is defined in \eqref{eq: latitude_weighting}.

\begin{equation}
L(j) = \frac{\cos(\text{lat}(j))}{\sum_{j=1}^{N_{\text{lat}}} \cos(\text{lat}(j))\label{eq: latitude_weighting}}
\end{equation}

 Model forecasts were also evaluated using the mean-latitude-weighted anomaly correlation coefficient (ACC). This metric ranges from -1 to 1, where a higher ACC value indicates better agreement between the model's predictions and the actual observations, while a lower ACC value indicates poorer agreement. The ACC assesses the consistency of spatial variations, regardless of their magnitudes. 

\begin{equation}
\text{ACC} = \label{eq: ACC} \frac{\sum_{i,j,k} L(j) \, \text{forecast}'_{i,j,k} \ \text{truth}'_{i,j,k}}{\sqrt{\sum_{i,j,k} L(j) \, \text{forecast}'^2_{i,j,k} {\sum_{i,j,k} L(j) \, \text{truth}'^2_{i,j,k}}}}
\end{equation}

In equation \eqref{eq: ACC}, \(forecast'\)  and \(truth'\) denote the difference to the climatology. We used the climatology as defined in section \ref{sec: Data Description}.

The Heidke Skill Score (HSS), as defined by NOAA, is used to compare the DUNE AI inferences with the NOAA operational forecasts. To evaluate forecast accuracy using the Heidke Skill Score (HSS), a contingency table is employed to categorize forecast outcomes into hits (H), misses (M), false alarms (F), and correct negatives (C). Hits (H) represent correct forecasts where predicted events match observed events. The total number of valid forecast-observation pairs (T) is the sum of all categories in the contingency table (H + M + F + C). More details on this can be found at \url{https://www.cawcr.gov.au/projects/verification/#Contingency_table}. The expected number of correct forecasts (E), assuming random chance, is typically one-third of T. The HSS is then calculated using the formula:

\begin{equation}
\label{eq: HSS}
\text{HSS} \% = 100 \times \frac{H - E}{T - E}
\end{equation}

In equation \eqref{eq: HSS}, \(H\) denotes the number of correct forecasts, \(E\) denotes the expected number of correct forecasts (1/3 of total), and \(T\) is the total number of valid forecast-observation pairs.

 The HSS measures how often the forecast category correctly matches the observed category above the number of correct hits expected by chance alone. HSS utilizes the number of correct and incorrect category hits, with values ranging from -50 to 100. A score of 100 indicates a perfect forecast, while a score of -50 indicates a perfectly incorrect forecast. Scores greater than 0 indicate improvement compared to a random forecast, thus indicating skill \citep{NOAA_HSS}.

We categorized the data using percentiles based on the climatology (1991-2020). For any given month, if the inference/observation value is less than the lowest 33\textsuperscript{rd} percentile, it is classified as "below normal"; if it is higher than the 66\textsuperscript{th} percentile, it is classified as "above normal"; and if it falls between the 33\textsuperscript{rd} and 66\textsuperscript{th} percentiles, it is categorized as "near normal." The results with the Heidke Skill Score (HSS) are in Section \ref{sec: Comparison with NOAA's Monthly Forecasts}.

\section{Methodology}
\label{sec: Methodology}
The Deep UNet++ Ensemble (DUNE) AI forecast model is a specific encoder-decoder architecture based on the UNet++ architecture \citep{zhou2018unet++} proposed initially for medical image segmentation tasks. The UNet++ architecture is a variation of the famous U-Net architecture \citep{ronneberger2015u}, which is also a variation of the traditional encoder-decoder networks \citep{baldi2012autoencoders} that have succeeded in image segmentation tasks.  \citet{larraondo2019data} investigated the application of several CNNs, such as SegNet, VGG16, and U-Net, to predict precipitation fields from geopotential height fields using the reanalysis data and found that U-Net delivered the best forecasts.

\begin{figure}[hbtp]
  \centering
  \includegraphics[width=1\textwidth]{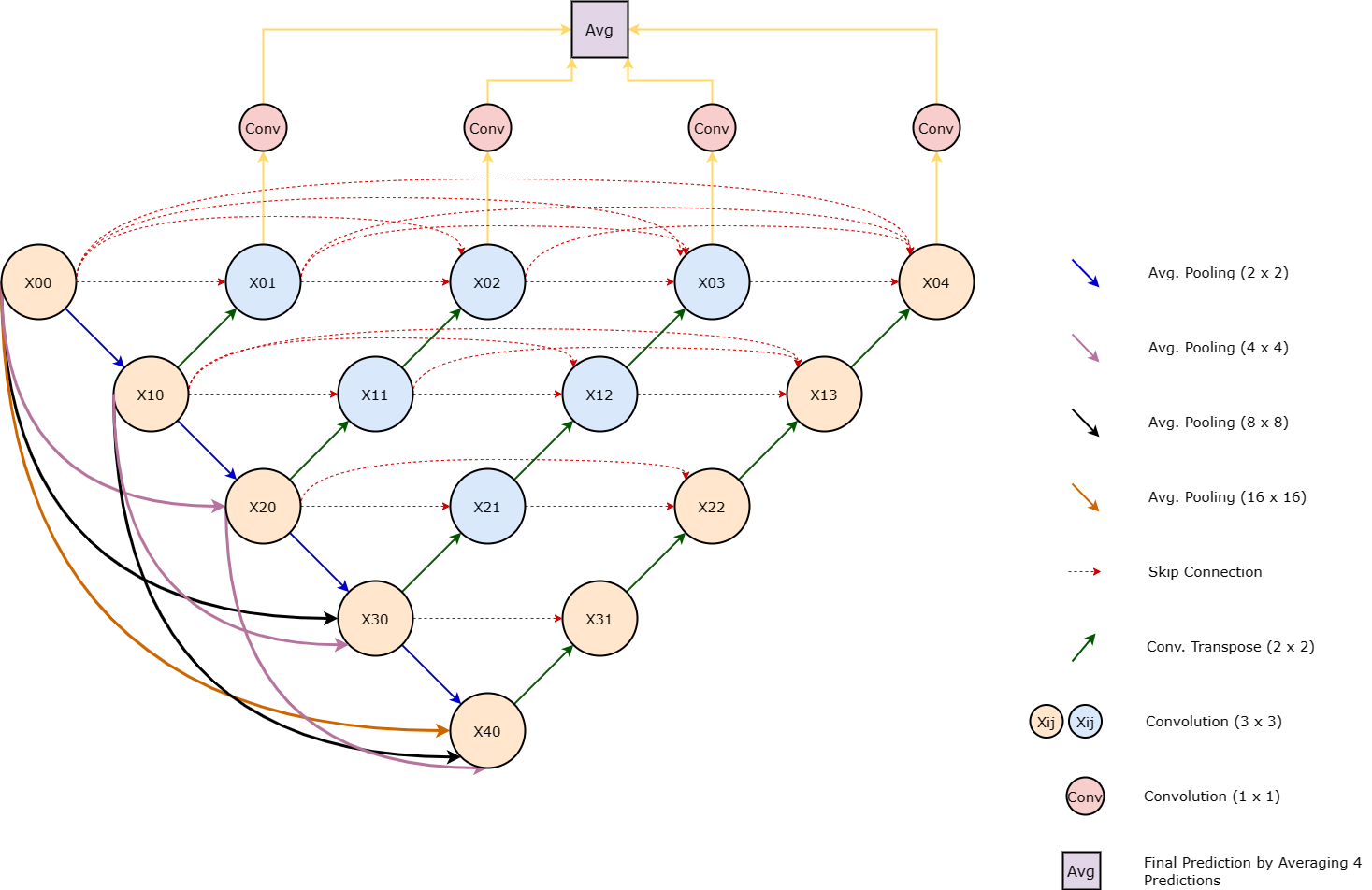}
  \caption{The DUNE AI Architecture}
  \label{fig:dune_ai}
\end{figure}

The DUNE AI forecast model, illustrated in Figure \ref{fig:dune_ai}, uses a fully convolutional neural network (FCN) \citep{long2015fully} to predict the atmospheric state from one monthly mean to the next. The convolution operations within the CNN learn a specified number of filters, each with a shape of M x N. These filters are translated across the input atmospheric state, producing an output state for each M x N input image multiplied by the filters. The expanded architecture is illustrated in Figure \ref{fig: DUNE}.

The initial weights for these filters are generated using the Kaiming initialization method \citep{he2015delving}, and the model learns their weights through gradient back-propagation during the training phase \citep{towardsaiGradientDescent}. These filters extract complex patterns and features from the input states, and their weights are shared across the input domain. This design allows the DUNE AI model to perform pixel-wise semantic calculation, enabling the generation of quantitative estimates of meteorological parameters such as T2m and SST at each latitude-longitude grid point.

Notably, the DUNE AI model has a relatively low number of trainable parameters, approximately 45,843,720. In contrast, a neural network that fully connects all input features to all output features would have a trillion parameters, which is impractical.

The DUNE architecture was based on the UNet++ architecture for a prediction task. The architecture employs convolution operations to learn filter weights and to increase the number of channels. Throughout the network, the same filter size (3 x 3) is used with padding to maintain dimensionality at each level. Two residual CNN blocks are utilized at each level, with the model updating filter weights iteratively through back-propagation. As the number of channels progressively increases, this augmentation enables the network to capture and learn more complex patterns and features from the input data and to better generalize to unseen data.
 
These convolutional operations are followed by a spatial pooling operation, which reduces the dimensionality of the atmospheric state by a factor of 2. This reduction is achieved by calculating the average value within each 2 x 2 sub-grid. After the pooling layer, convolutional operations are again applied with the same filter size (3 x 3) on states with progressively coarser spatial resolution. Through this process, the CNNs learn filters that represent larger-scale atmospheric patterns. This sequence constitutes the contracting path, or the encoder part of the architecture, effectively summarizing the information by reducing the dimensionality of the atmospheric state while increasing the number of channels.

In a CNN, low-level feature maps at higher resolutions capture detailed texture information, such as fine-scale temperature variations and localized anomalies. This allows the model to detect subtle patterns and gradients in temperature distribution. In contrast, high-level feature maps at lower resolutions encode more abstract semantic information, capturing broader climatic trends and large-scale patterns such as seasonal cycles and long-term temperature changes. We often deepen the network to enhance our model's feature extraction ability. Inspired by the DenseNet architecture \citep{huang2017densely}, we implemented an extreme form of skip connections to supply low-level features to deeper network layers. Specifically, we passed each layer's features to all subsequent layers and then fused them through concatenation. This was achieved by applying pooling operations of varying sizes to the higher-dimensional feature maps before integrating them with the lower-dimensional ones. Including additional feature maps from higher dimensions to help the model capture more accurate and subtle patterns.

\begin{figure}[hbtp]
  \centering
  \includegraphics[width=1\textwidth]{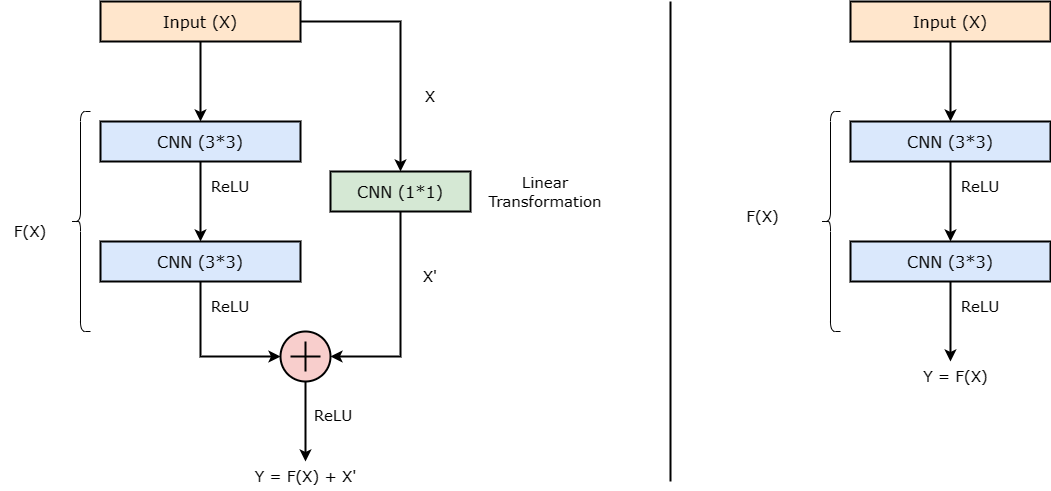}
  \caption{(a) Residual Block (b) Regular CNN Block}
  \label{fig: residual block}
\end{figure}

As the network gets deeper, one often faces the problem of vanishing gradients and network degradation \citep{hochreiter1998vanishing}. To avoid these problems and ensure a faster model convergence, we used a residual CNN block \citep{he2016deep} instead of a CNN block at each level. To create a residual block, we applied a linear transformation using a CNN with filter size 1 x 1 to the original input so that it has the same number of channels required for the summation operation at the end. The residual CNN block is shown in Figure \ref{fig: residual block}(a), and a regular CNN block is shown in Figure \ref{fig: residual block}(b).

In the expanding path or the decoder part of the architecture, transposed convolutions \cite{zeiler2010deconvolutional} are used to increase the dimensionality of the state of the atmosphere. At each up-sampling layer, the spatial dimensionality is increased by a factor of 2. This operation is continued until one reaches the dimensions of the original input.

However, down-sampling operations lead to the loss of spatial information, potentially resulting in overly smooth predictions. U-Net addresses this by concatenating the tensor state of the CNN at each encoding step with the tensor state immediately following each up-sampling operation in the decoding phase. These skip connections reduce information loss after the bottleneck layer and aid in recovering fine-grained details.

In UNet++, these skip connections are redesigned to minimize the semantic gap between the features incoming from the sub-networks of the encoder and the decoder, making learning more effective. By doing so, the network can learn more abstract and meaningful representations of the input data. Additionally, integrating additional convolutional layers further enhances the network's ability to extract and refine features, resulting in a significantly improved feature representation that benefits prediction tasks.

Four predictions were generated by trimming our architecture to various depths, using the power of ensemble forecasts within a single architecture. Next, convolutional layers are applied to each output with a filter size of 1 x 1 to consolidate multiple channel outputs into a single channel. Finally, a final prediction is created by averaging these four ensemble predictions from the four channels.

Except for the final four convolutions, which utilize a filter size of (1 x 1), each convolutional operation in the architecture is followed by a ReLU nonlinear activation function \cite{agarap2018deep}.

The model is trained using the Adam optimizer \cite{kingma2014adam}, a version of the stochastic gradient descent optimization algorithm \cite{loshchilov2016sgdr}, with a learning rate of 1e-3, a batch size of 4, and a weight decay of 1e-4. Additionally, the CosineAnnealing learning rate scheduler is used with a maximum of 225 iterations to adjust the learning rate. It uses the latest learning rate after the 225\textsuperscript{th} epochs. A mean latitude-weighted root-mean-squared-error is used as the loss function (see section \ref{sec: Evaluation}). The model is trained for 500 epochs to ensure effective loss minimization, with early stopping based on the validation set loss. If no new validation loss minimum is observed within 100 epochs during training, the training stops, and the model weights yielding the smallest validation loss are restored for further testing.

The model was constructed using the open-source PyTorch library \citep{paszke2019pytorch} from Meta AI in Python. For most of the data pre-processing pipeline, the open-source xarray project was utilized  \citep{hoyer2017xarray}. The model training was conducted on 8 Tesla A100-40 GB GPUs on the NASA Center for Climate Simulation (NCCS) servers at the Goddard Space Flight Center. The Matplotlib library \citep{Hunter:2007} was used for all visualizations, employing Basemap for map projections.

\textbf{Long-Range Forecast using the Moving Window Approach:} The DUNE AI model can forecast the monthly mean, seasonal mean, or annual mean temperatures one monthly/seasonal/annual mean in advance. For example, given the prior month’s monthly mean temperature, it will forecast the next month’s monthly mean temperature. Inspired by \citep{tripura2023wavelet}, a moving window approach was used to extend the model’s capabilities to forecast more than one month in advance. 

In the moving window approach, the prior x months are used to forecast the next x months. For example, if one wants to forecast the temperature for the next 12 months, the prior 12 months’ temperatures and constants can be used. Note that there are 12 inputs for TISR and 5 channels for the other constants. The results of DUNE are presented in section \ref{sec: Monthly Mean Extended-Range Forecast Results}. More information on this can be found in the Appendix.

\section{DUNE Forecast Results}
\label{sec:DUNE Forecast Results}
This section presents the monthly, seasonal, and annual mean forecast experiments using the DUNE AI method. These forecasts are compared to three other baseline methods: persistence, climatology, and multiple linear regression. Regional performances, derived from the global forecast, are separately analyzed with respect to performance statistics. 

\subsection{Global Forecast Comparison: DUNE AI versus Other AI Methods}
Table \ref{tab:comparison_table} presents a comparative analysis of the DUNE AI method against other AI-based forecasting approaches using monthly data. The results indicate that when trained on monthly mean data, the DUNE AI model yields superior global forecast accuracy compared to other AI models that utilize hourly or daily data for monthly forecasting (shown in table \ref{tab:comparison_table}). Specifically, the DUNE AI method demonstrates a twofold reduction in RMSE relative to other AI methods when employing monthly mean data as input.

\begin{table}[H]
\centering
\resizebox{\textwidth}{!}{
\begin{tabular}{ccccccc}
\toprule
Evaluation Method & Forecast Time & Climax (5.625\textdegree) & Climax (1.40625\textdegree) & ResNet (5.625\textdegree) & UNet (5.625\textdegree) & DUNE AI (0.25\textdegree)\\
\midrule
RMSE\textsuperscript{a} ($\downarrow$) & 720hrs & 2.67 & 2.74 & 2.86 & 2.81 & \textbf{1.07}\textsuperscript{c}\\
ACC\textsuperscript{b} ($\uparrow$)  & 720hrs & 0.85 & 0.84 & 0.83 & 0.83 & \textbf{0.95/0.74}\textsuperscript{b, c}\\
\bottomrule
\end{tabular}
}
\caption{Comparison of DUNE AI monthly forecasts (bold) with other AI hourly/daily averaged methods}\label{tab:comparison_table}
\footnotesize{$^{\rm a}$ The DUNE AI method generates monthly mean forecasts, whereas the other methods in the table produce hourly forecasts up to 720 hours in advance. This highlights the advantage of using monthly mean data.\\
$^{\rm b}$ To calculate the ACC, the DUNE AI method used climatology data from 1950 to 1979. In contrast, the other AI methods in the table used climatology data from their respective test periods. If the DUNE AI method used the climatology from the test period, it gives the ACC of 0.95, as shown in the table. It is worth noting that the DUNE AI forecast is at a significantly finer resolution. The table values are adopted from \citep{nguyen2023climax}.
$^{\rm c}$ Results using monthly mean data.}
\end{table}

\subsection{Global Monthly Mean Forecasts}
\label{sec: Global Monthly Mean Forecasts}
This experiment uses the inferred weights computed from training the DUNE AI model to use the prior month to forecast the successive monthly mean temperatures using the ERA5 input data from 1980 to 2016. Given the temperature anomalies and the six constants, DUNE infers the temperature anomalies for the next month. In particular, using data from 1980 to 2016 (444 months) for training successive months, and then data from the years 2017 and 2018 (24 months) to validate the forecast inference, forecasts (i.e., inferences) are then made employing the derived weights to infer monthly data from all months between 2019 and 2023 (60 months). The wall clock time required for processing and inferring forecasts over 60 months was approximately 20 seconds.

Table \ref{tab:global_monthly_table} compares the DUNE AI model's forecast skills (inferences) with the baselines, persistence, climatology, and multiple linear regression. The average of RMSE, ACC, and HSS is presented in table \ref{tab:global_monthly_table} over the testing period from 2019 to 2023 (5*12 = 60 months). The DUNE AI method outperforms all the baselines for monthly mean forecasts. We utilized the Scikit-learn library \citep{scikit-learn} to calculate statistics for the multiple linear regression baseline. Since multiple linear regression consistently yields poorer performance than other baselines, they were not included in further regional results.

\begin{table}[hbtp]
\centering
\resizebox{\textwidth}{!}{%
\begin{tabular}{lcccccccccccccccccc}
\toprule
Method & \multicolumn{3}{c}{Global} & \multicolumn{3}{c}{Global Land} & \multicolumn{3}{c}{Global Oceans} & \multicolumn{3}{c}{US} & \multicolumn{3}{c}{Australia} & \multicolumn{3}{c}{Boreal Forests} \\
\cmidrule(r){2-4} \cmidrule(lr){5-7} \cmidrule(lr){8-10} \cmidrule(lr){11-13} \cmidrule(lr){14-16} \cmidrule(lr){17-19}
 & RMSE ($\downarrow$) & ACC ($\uparrow$) & HSS ($\uparrow$) & RMSE ($\downarrow$) & ACC ($\uparrow$) & HSS ($\uparrow$) & RMSE ($\downarrow$)& ACC ($\uparrow$) & HSS ($\uparrow$) & RMSE ($\downarrow$)& ACC ($\uparrow$) & HSS ($\uparrow$) & RMSE ($\downarrow$)& ACC ($\uparrow$) & HSS ($\uparrow$) & RMSE ($\downarrow$)& ACC ($\uparrow$) & HSS ($\uparrow$)\\
\midrule
Persistence (Prior Month) & 2.76 & 0.33 & 19.39 & 4.41 & 0.27 & 11.46 & 1.16 & 0.59 & 23.46 & 4.68 & 0.19 & 13.01 & 2.96 & 0.13 & 5.87 & 6.82 & 0.16 & 5.38 \\
Persistence (Prior Year Same Month) & 1.52 & 0.53 & 9.71 & 2.28 & 0.51 & 7.16 & 0.90 & 0.61 & 11.02 & 2.39 & 0.42 & 5.54 & 1.82 & 0.22 & 1.15 & 3.25 & 0.46 & 4.36 \\
Climatology (1950-1979) & 1.60 & 0  &-11.24 & 2.38 & 0 & -12.30 & 1.03 & 0 & -10.70 & 2.26 & 0 &-10.40 & 1.51 & 0 &-5.20 & 3.18 & 0 & -9.10\\
Multiple Linear Regression (2\textdegree resolution)  & 1.96 & 0.42 & - & - & - & - & - & - & - & - & - & - \\
DUNE AI Method & 1.07 & 0.74  & 26.84 & 1.65 & 0.69 & 10.95 & 0.52 & 0.85 & 34.95 & 1.80 & 0.59 & 9.65 & 1.32 & 0.45 & 8.13 & 2.41 & 0.65 & 18.71\\
\bottomrule
\end{tabular}%
}
\caption{Comparison of five-year average (2019-2023) monthly mean forecasts using RMSE, ACC, and HSS metrics. The results demonstrate that DUNE AI inferences significantly improve performance compared to baseline models, including multiple linear regression. The DUNE AI model outperforms the baselines in all the regions.}
\label{tab:global_monthly_table}
\end{table}

\begin{figure}[hbtp]
  \centering
  \includegraphics[width=1\textwidth]{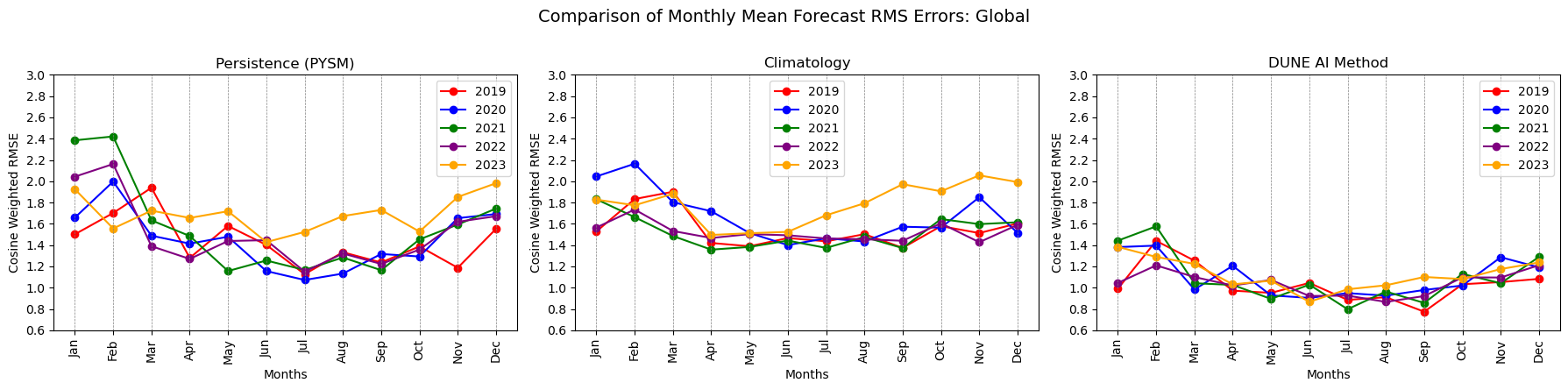}
  
\caption{Global RMSE of Monthly Mean Forecasts: (a) Persistence (PYSM), (b) Climatology, and (c) DUNE AI Method. The comparison shows that the DUNE AI model consistently achieves lower RMSE values, indicating superior forecasting skills. Notably, during the record warm year of 2023, the DUNE AI model exhibited relatively small variability compared to the baseline models. Over the five years, the DUNE AI model displays slightly higher variance in January, February, and December but maintains a uniformly small variance for the remaining months.}
  \label{fig:monthly_rms_global}
\end{figure}

\begin{figure}[hbtp]
  \centering
  \includegraphics[width=1\textwidth]{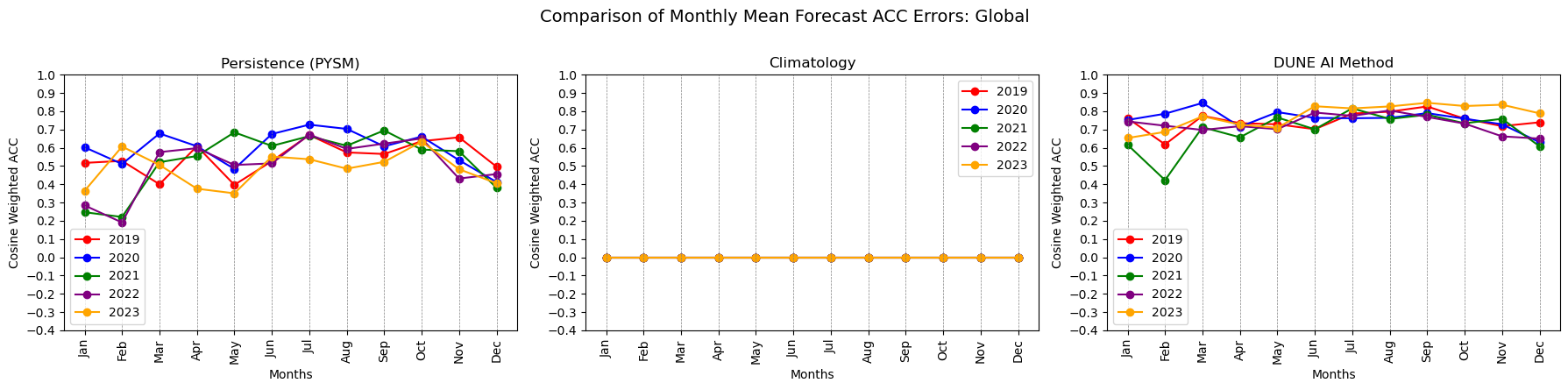}
\caption{Global ACC of Monthly Mean Forecasts: (a) Persistence (PYSM), (b) Climatology, and (c) DUNE AI Method. The comparison shows that the DUNE AI model consistently achieves higher ACC values, indicating superior forecasting skills. Notably, during the record warm year of 2023, the DUNE AI model exhibited relatively small variability compared to the baseline models. Over the five years, the DUNE AI model displays higher ACC value of 0.8 compared to 0.5 (baselines) in addition to maintaining a smaller variance.}
  \label{fig:monthly_acc_global}
\end{figure}

Figure \ref{fig:monthly_rms_global} and Figure \ref{fig:monthly_acc_global} show the global RMSE and global ACC for the test period from 2019 to 2023, respectively. For RMSE, the lower, the better; for ACC, the higher, the better. The DUNE AI model generated better forecasts for all the months compared to the baselines. 

\begin{figure}[H]
  \centering
  \includegraphics[width=1\textwidth]{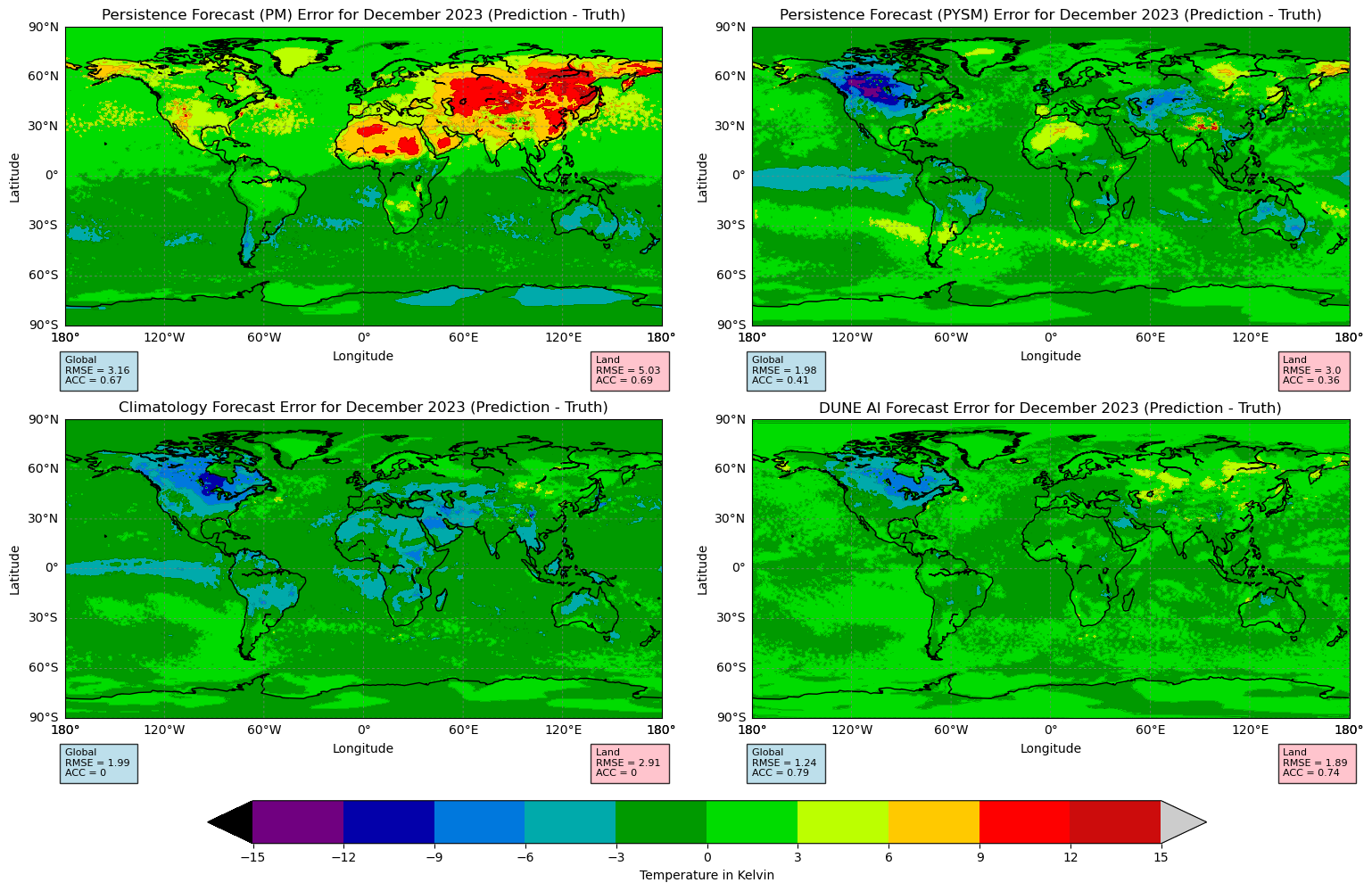}
  \caption{Visualization of forecast errors for December 2023: (a) Persistence (PM), (b) Persistence (PYSM), (c) Climatology, (d) DUNE AI Method. RMSE and ACC values are displayed below each plot. (a) and (b) highlight significant prediction errors using the Persistence method over North America and Eurasia. In contrast, the DUNE AI model demonstrates substantial performance improvements, particularly in the El Niño region, South America, Africa, North America, and Boreal regions across North America and Eurasia.}
  \label{fig:monthly_global_december_2023}
\end{figure}

Since December 2023 exhibits the highest temperature variability, Figure \ref{fig:monthly_global_december_2023} compares forecast errors (prediction-truth) from the DUNE AI model with baseline methods for this month. In the figure, forecast errors closer to zero (depicted in green) indicate improved forecast accuracy. The visual assessment demonstrates that the DUNE AI model's forecast errors are substantially lower than those from the baseline methods.

\begin{figure}[H]
  \centering
  \includegraphics[width=1\textwidth]{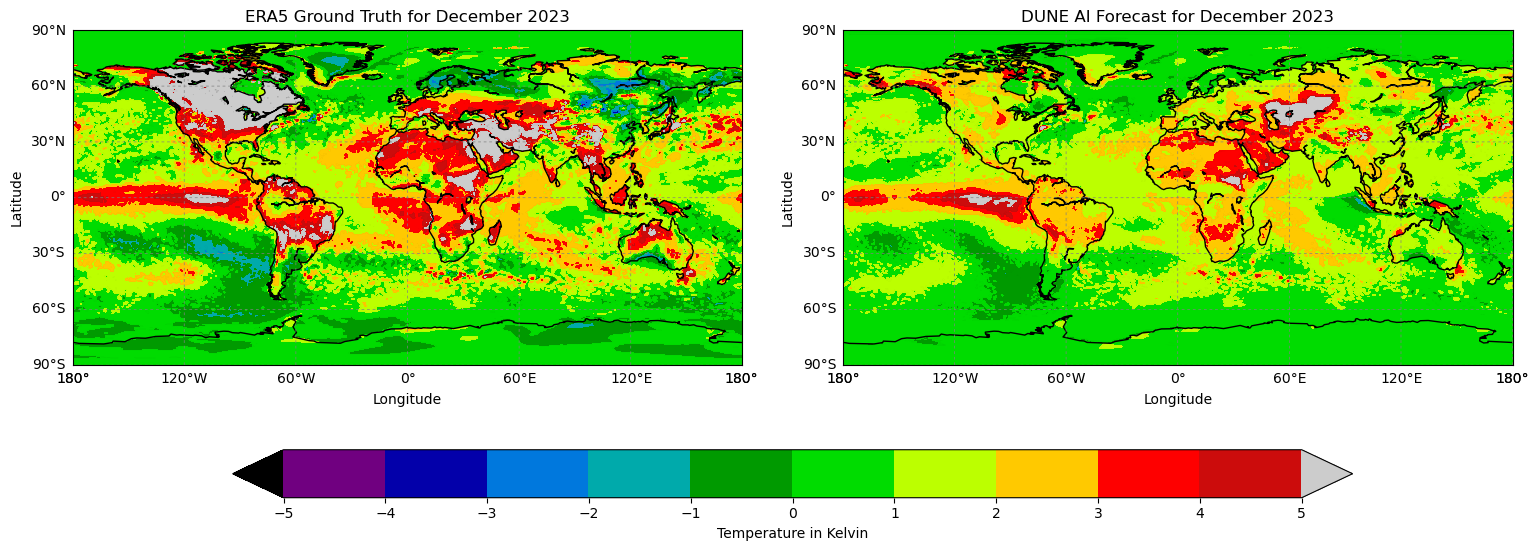}
  \caption{A visual comparison of ERA5 ground truth anomalies and the DUNE AI forecast anomalies for December 2023: (a) ERA5 ground truth anomalies for December 2023, (b) DUNE AI forecast anomalies for December 2023. This comparison illustrates the alignment between the observed data and the AI model's predictions, highlighting the DUNE AI's capability to replicate the observed anomalies for December 2023 accurately. The DUNE AI model notably captured anomalies in the El Niño region near South America. It also successfully predicted anomalies in North America, though these predictions were less pronounced than the observed anomalies. Additionally, the model accurately predicted the Western Australian Current and the Equatorial Countercurrent. However, the DUNE AI model underpredicted anomalies over Australia and failed to capture the lower anomalies in the Amazon Rainforest.}
  \label{fig:global_december_2023_forecast}
\end{figure}

Figure \ref{fig:global_december_2023_forecast} compares ERA5 ground truth anomalies with the DUNE AI forecast anomalies for December 2023, which shows the DUNE AI forecast is very close to the ERA5 ground truth anomalies. 

\subsection{Global Seasonal Mean Forecasts}
\label{sec: Global Seasonal Mean Forecasts}
The following experiments were designed to predict seasonal mean temperatures one season in advance. A season is defined as the average of the temperatures from the three months comprising one of the four seasons. Using data from January 1980 to December 2016, the first season available for prediction is Summer 1980, as Spring would require data from December 1979. The last season available for prediction is Fall 2016, as Winter would require data from January 2017. Consequently, for the 36 years of training data, there were 142 seasons available for training seasonal predictions, 6 seasons for validation, and 18 seasons for testing.

Model prediction statistics were computed for the following baselines: Persistence using Prior Season (PS), Persistence using Prior Year Same Season (PYSS), and Climatology from 1950-1979. The table shows the average RMSE and ACC over 18 seasons from 2019 to 2023. Table \ref{tab: global_seasonal_table} presents global forecast comparisons. Table \ref{tab: global_seasonal_table} shows the average RMSE and ACC over the testing period from 2019 to 2023. Some supplementary graphs and visualizations for the seasonal mean experiment are in the Appendix. 

Improved results for seasonal forecasts can be attributed to the reduced variability in seasonal mean temperatures compared to monthly means.

\begin{table}[hbtp]
\centering
\resizebox{\textwidth}{!}{%
\begin{tabular}{lcccccccccccc}
\toprule
Method & \multicolumn{2}{c}{Global} & \multicolumn{2}{c}{Global Land} & \multicolumn{2}{c}{Global Oceans} & \multicolumn{2}{c}{USA} & \multicolumn{2}{c}{Australia} & \multicolumn{2}{c}{Boreal Forests} \\
\cmidrule(r){2-3} \cmidrule(lr){4-5} \cmidrule(lr){6-7} \cmidrule(lr){8-9} \cmidrule(lr){10-11} \cmidrule(lr){12-13}
 & RMSE ($\downarrow$) & ACC ($\uparrow$) & RMSE ($\downarrow$) & ACC ($\uparrow$) & RMSE ($\downarrow$) & ACC ($\uparrow$) & RMSE ($\downarrow$) & ACC ($\uparrow$) & RMSE ($\downarrow$) & ACC ($\uparrow$) & RMSE ($\downarrow$) & ACC ($\uparrow$)\\
\midrule
Persistence (Prior Season)  & 6.24 & 0.12 & 10.12 & 0.09 & 2.26 & 0.35 & 11.18 & 0.005 & 6.74 & 0.09 & 16.34 & -0.006 \\
Persistence (Prior Year Same Season) & 1.09 & 0.67 & 1.52 & 0.69 & 0.76 & 0.65 & 1.35 & 0.22 & 1.35 & 0.33 & 2.18 & 0.67 \\
Climatology (1950-1979) & 1.37 & 0 & 1.94 & 0 & 0.95 & 0 & 1.65 & 0 & 1.23 & 0 & 2.57 & 0 \\
Multiple Linear Regression (2\textdegree resolution)  & 1.65 & 0.48 & - & - & - & - & - & - & - & - & - & - \\
DUNE AI Method & 0.87 & 0.77 &  1.20 & 0.79 & 0.63 & 0.74 & 1.12 & 0.79 & 1.10 & 0.44 & 1.66 & 0.77 \\
\bottomrule
\end{tabular}%
}
\caption{Comparison of five-year average (2019-2023) seasonal mean forecasts using RMSE and ACC metrics. The results demonstrate that DUNE AI inferences significantly improve performance compared to baseline models, including multiple linear regression. The DUNE AI model outperforms the baselines in all the regions.}
\label{tab: global_seasonal_table}
\end{table}

\subsection{Global Yearly Mean Forecasts}
\label{sec:Global Yearly Mean Forecasts}
Similarly, predicting yearly mean temperatures a year in advance leaves 37 years for training, 2 years for validation, and 5 years for testing. One could increase the number of seasons and years for predictions by creating sliding 3-month seasons and years different from the calendar, but this was not done for this study.  Table \ref{tab:global_yearly_table} compares global yearly mean RMSE and ACC results. Visualization results are in the Appendix.

Enhanced accuracy in yearly mean forecasts is attributed to the reduced variability in annual mean temperatures compared to seasonal or monthly means.

\begin{table}[hbtp]
\centering
\resizebox{\textwidth}{!}{%
\begin{tabular}{lcccccccccccc}
\toprule
Method & \multicolumn{2}{c}{Global} & \multicolumn{2}{c}{Global Land} & \multicolumn{2}{c}{Global Oceans} & \multicolumn{2}{c}{USA} & \multicolumn{2}{c}{Australia} & \multicolumn{2}{c}{Boreal Forests} \\
\cmidrule(r){2-3} \cmidrule(lr){4-5} \cmidrule(lr){6-7} \cmidrule(lr){8-9} \cmidrule(lr){10-11} \cmidrule(lr){12-13}
 & RMSE ($\downarrow$) & ACC ($\uparrow$) & RMSE ($\downarrow$) & ACC ($\uparrow$) & RMSE ($\downarrow$) & ACC ($\uparrow$) & RMSE ($\downarrow$) & ACC ($\uparrow$) & RMSE ($\downarrow$) & ACC ($\uparrow$) & RMSE ($\downarrow$) & ACC ($\uparrow$)\\
\midrule
Persistence &0.67 & 0.83 & 0.86 & 0.86 & 0.52 & 0.78 & 0.91 & 0.83 & 0.76 & 0.64 & 1.33 & 0.84 \\
Climatology (1950-1979) &1.20 & 0 & 1.66 & 0 & 0.83 & 0 & 1.49 & 0 & 0.90 & 0 & 2.25 & 0 \\
Multiple Linear Regression (2\textdegree resolution)  & 1.40 & 0.55 & - & - & - & - & - & - & - & - & - & - \\
DUNE AI Method & 0.63 & 0.85 & 0.77 & 0.89 & 0.53 & 0.77 & 0.79 & 0.85 & 0.74 & 0.64 & 1.08 & 0.89 \\
\bottomrule
\end{tabular}%
}
\caption{Comparison of five-year average (2019-2023) annual mean forecasts using RMSE and ACC metrics. The results demonstrate that DUNE AI inferences significantly improve performance compared to baseline models, including multiple linear regression. The DUNE AI model outperforms the baselines in all the regions.}
\label{tab:global_yearly_table}
\end{table}

\subsection{Validation and Testing Period Statistics Comparison}

\begin{table}[hbtp]
\centering
\resizebox{\textwidth}{!}{%
\begin{tabular}{lcccccccccccc}
\toprule
DUNE AI Forecast & \multicolumn{2}{c}{Monthly Mean} & \multicolumn{2}{c}{Seasonal Mean} & \multicolumn{2}{c}{Annual Mean} \\
\cmidrule(r){2-3} \cmidrule(lr){4-5} \cmidrule(lr){6-7} 
 & RMSE ($\downarrow$) & ACC ($\uparrow$) & RMSE ($\downarrow$) & ACC ($\uparrow$) & RMSE ($\downarrow$) & ACC ($\uparrow$) &\\
\midrule
Validation Period (2017-2018)  & 1.03 & 0.71 & 0.80 & 0.76 & 0.50 & 0.87  \\
Testing Period (2019-2023) & 1.07 & 0.74 & 0.87 & 0.77 & 0.63 & 0.85\\
\bottomrule
\end{tabular}%
}
\caption{Comparison of DUNE AI monthly/seasonal/annual mean forecasts for validation (2017-2018) and testing (2019-2023) period.}
\label{tab: global_comparison_table}
\end{table}

\begin{figure}[H]
  \centering
  \includegraphics[width=0.5\textwidth]{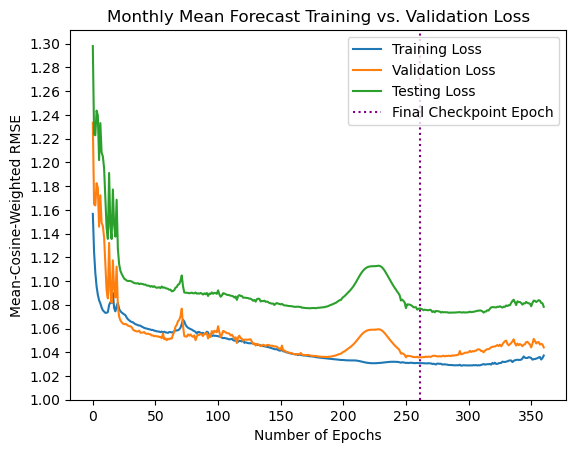}
  \caption{Monthly Mean Forecast Training, Validation, and Testing loss}
  \label{fig:monthly_mean_loss}
\end{figure}

Figure \ref{fig:monthly_mean_loss} shows the training, validation, and testing loss.

\subsection{Monthly Mean Extended-Range Forecast Results}
\label{sec: Monthly Mean Extended-Range Forecast Results}
In this section, DUNE monthly mean forecast results are presented using the moving window approach for extended periods. Using this approach, monthly mean forecasts were conducted for 1, 2, 3, 4, 6, and 12 months in advance. 

Table \ref{tab: global_moving_window} compares global monthly mean forecast results. Although the forecast skills of the DUNE model decrease with an extended forecast period, as anticipated, they remain superior to the baseline methods. Additionally, the RMSE and ACC plots are provided in the appendix for further reference.

\begin{table}[hbtp]
\centering
\resizebox{\textwidth}{!}{%
\begin{tabular}{lccccc}
\toprule
& & \multicolumn{2}{c}{Global} & \multicolumn{2}{c}{Global Land} \\
\cmidrule(lr){3-4} \cmidrule(lr){5-6}
Moving Window Frame & Method & RMSE  ($\downarrow$) & ACC ($\uparrow$) & RMSE   ($\downarrow$) & ACC ($\uparrow$) \\
\midrule
- & Persistence (Prior Month) & 2.76 & 0.33 & 4.30 & 0.27 \\
- & Persistence (Prior Year Same Month) & 1.52 & 0.53 & 2.23 & 0.51 \\
- & Climatology (1950-1979) & 1.60 & 0 & 2.29 & 0 \\
- & Multiple Linear Regression (2\textdegree resolution) & 1.96 & 0.42 & - & -\\
One Month & DUNE AI Method & 1.07 & 0.74 & 1.65 & 0.69 \\
Two Months & DUNE AI Method & 1.13 & 0.70 & 1.66 & 0.68 \\
Three Months & DUNE AI Method & 1.16 & 0.69 & 1.67 & 0.68 \\
Four Months & DUNE AI Method & 1.17 & 0.68 & 1.68 & 0.67 \\
Six Months & DUNE AI Method & 1.20 & 0.66 & 1.69 & 0.67 \\
Twelve Months & DUNE AI Method & 1.27 & 0.61 & 1.75 & 0.64 \\
\bottomrule
\end{tabular}%
}
\caption{A five-year (2019-2023) comparison of average monthly mean forecasts using the Moving Window Approach reveals an expected increase in error statistics and a decrease in forecasting skills as the moving window extends from 1 month to 12 months. Notably, the decline in skill over land is minimal and remains superior to all baseline results.}\label{tab: global_moving_window}
\end{table}
\subsection{Comparison with NOAA's Monthly Forecasts}
\label{sec: Comparison with NOAA's Monthly Forecasts}

This section provides a comparative analysis between the forecasts generated by the DUNE AI model and those issued by the National Oceanic and Atmospheric Administration (NOAA). The NOAA forecasts are designed explicitly for the United States, whereas the DUNE AI model is trained and inferred from training to predict T2m globally. For a direct comparison, statistics for the United States were extracted from the global output of the DUNE AI model. Notably, NOAA's forecasts are produced at a spatial resolution of 2\textdegree x 2\textdegree, whereas the DUNE AI model provides inferences at a significantly finer resolution of 0.25\textdegree x 0.25\textdegree. The NOAA Climate Prediction Center Gridded Monthly Forecasts and an explanation of their products can be accessed at: \url{https://www.cpc.ncep.noaa.gov/products/predictions/long_range/tools/briefing/mon_veri.grid.php}.

For this experiment, forecast and observation categories (i.e., above, near, and below normal as defined by NOAA ) were based on the percentile values of the climatology (1991-2020). For each month, if the forecast or observation value falls below the 33\textsuperscript{rd} percentile of the climatology for that month, it is classified as below normal (indicated in blue). If the value exceeds the 66\textsuperscript{th} percentile, it is classified as above normal (shown in red). Values between the 33\textsuperscript{rd} and 66\textsuperscript{th} percentiles are classified as near normal. These classifications for the DUNE forecasts have been validated by comparing ERA5 observations. NOAA categories are validated by NOAA relative to the NOAA observations and taken from the above NOAA Climate Center link. The climatology and categories used for the DUNE results are intended to be consistent with those described by the NOAA CPC\citep{NOAA_Blogs}.

\begin{figure}[H]
\centering
\includegraphics[width=1\textwidth]{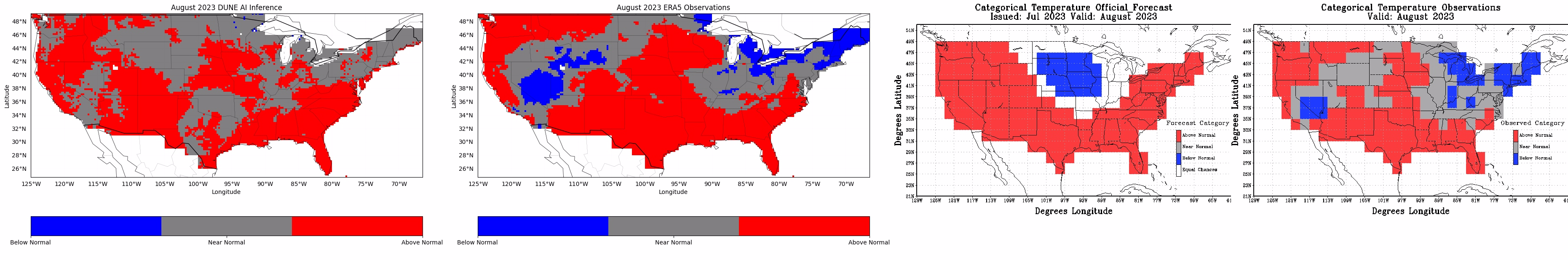}
\caption{Comparison of forecasts and observations for August 2023: (a) DUNE AI Inference with an HSS of 34.18, (b) ERA5 Observations, (c) NOAA Operational Forecast with an HSS of 23.49, (d) NOAA Observations. The DUNE AI model successfully inferred above-normal temperatures on the west coast, midwest, and southeast coast. Additionally, in the northeast, the DUNE AI model predicted near-normal temperatures, while NOAA forecasted higher temperatures. However, the actual temperatures in the northeast were below normal. Overall, the DUNE AI model performed very well, demonstrating superior performance compared to the NOAA forecast.}
\label{fig: August_2023}
\end{figure}

The following figures selectively show monthly forecast comparisons of DUNE and the CPC methodology to illustrate the difference between observations at 0.25 and 2.0 degrees in performing HSS accuracy. Figure \ref{fig: August_2023} provides a comparative analysis of the DUNE AI inferences and NOAA official forecasts relative to observations for August 2023. The analysis reveals that the DUNE AI inference, with a Heidke Skill Score (HSS) of 34.18, substantially outperformed the NOAA official forecast for this case, which has an HSS of 23.49, as the NOAA measure of predictive accuracy.

\begin{figure}[H]
\centering
\includegraphics[width=1\textwidth]{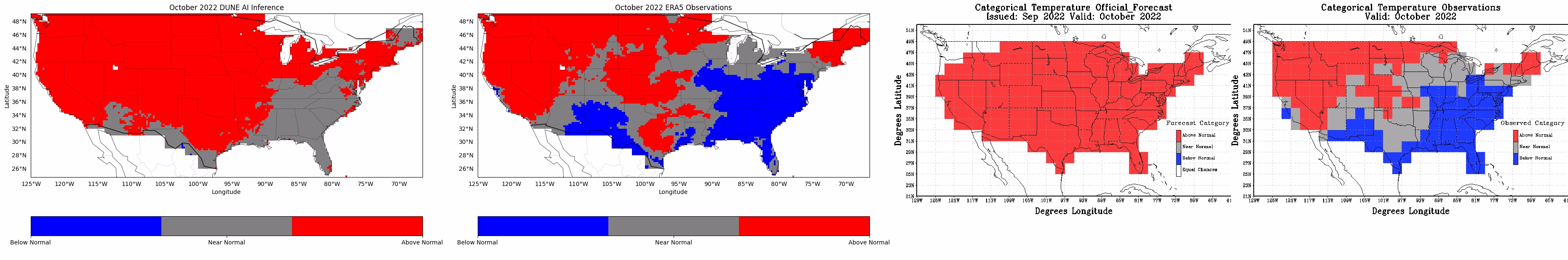}
\caption{Comparison of forecasts and observations for October 2022: (a) DUNE AI Inference with an HSS of 29.97, (b) ERA5 Observations, (c) NOAA Operational Forecast with an HSS of 7.33, (d) NOAA Observations. The DUNE AI model successfully captured the higher temperature anomalies in the western region of the United States. However, in the southeastern region, it failed to capture the below-normal anomalies. Comparing the DUNE AI and NOAA forecasts in the southeastern region, the DUNE AI model performed comparatively better. It predicted near-normal temperatures, whereas the NOAA forecast predicted above-normal temperatures, even though the observed values were below normal.}
\label{fig: October_2022}
\end{figure}

Figure \ref{fig: October_2022} presents a comparative analysis of forecast and observation data for October 2022. The DUNE AI inference achieved a Heidke Skill Score (HSS) of 29.97, whereas the NOAA operational forecast had an HSS of 7.33. This analysis highlights the performance of the DUNE AI model compared to the NOAA forecasts for this month.

\begin{figure}[H]
\centering
\includegraphics[width=1\textwidth]{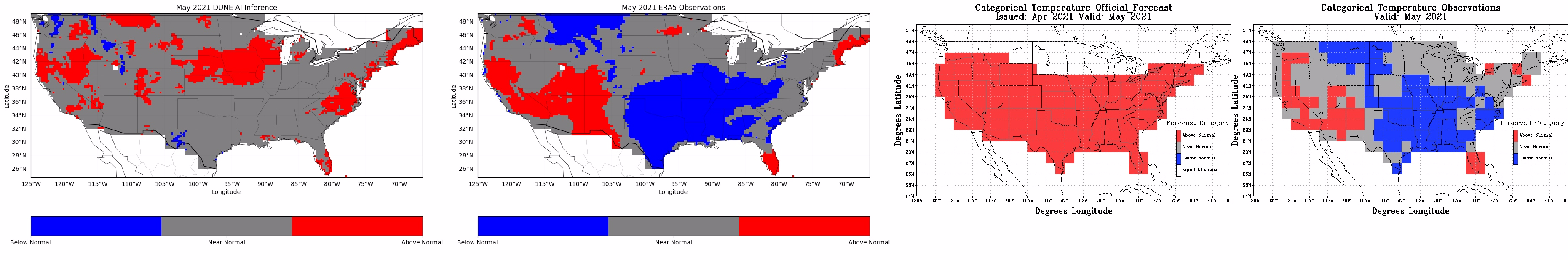}
\caption{Comparison of forecasts and observations for May 2021: (a) DUNE AI Inference with an HSS of 15.71, (b) ERA5 Observations, (c) NOAA Operational Forecast with an HSS of -16.59, (d) NOAA Observations. In May 2021, both DUNE AI and NOAA performed poorly in temperature forecasting. However, the DUNE AI model managed to capture above-normal temperatures in Florida, the northeastern coast, and parts of the western coast. Notably, in the central region, where below-normal temperatures were observed, the DUNE AI model inferred near-normal temperatures, while NOAA predicted above-normal temperatures.}

\label{fig: May_2021}
\end{figure}

Figure \ref{fig: May_2021} presents a comparative analysis of forecast and observation data for May 2021. The DUNE AI inference achieved a Heidke Skill Score (HSS) of 15.71, whereas the NOAA operational forecast had an HSS of -16.59. This analysis highlights the performance of the DUNE AI model compared to the NOAA forecasts for this month.

\begin{figure}[H]
  \centering
  \includegraphics[width=0.8\textwidth]{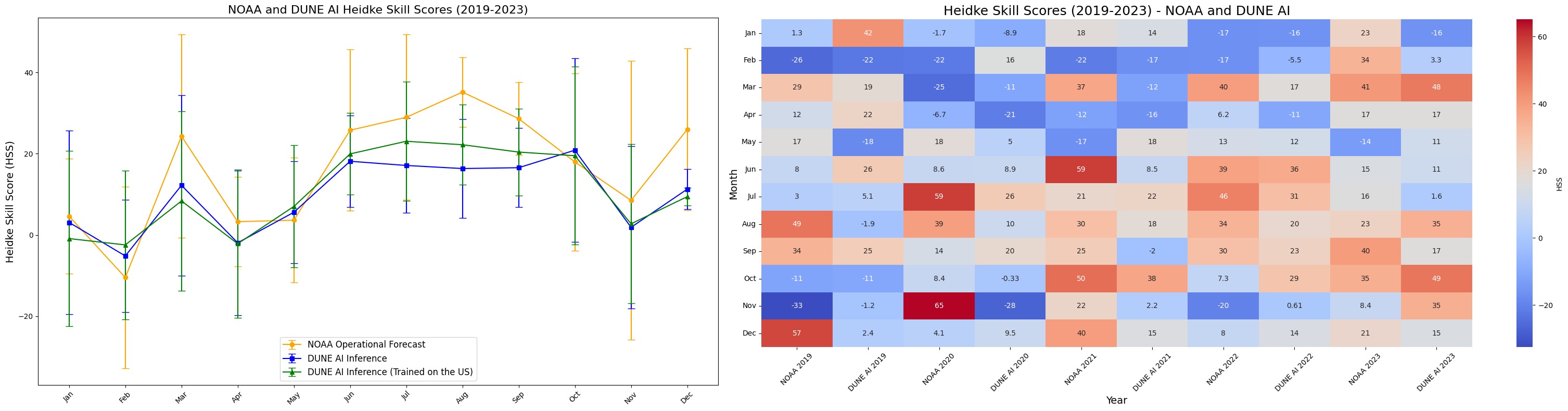}
  \caption{Heidke Skill Score (HSS) Comparison for Monthly Averaged Forecasts (2019-2023) (a) Comparison of monthly averaged HSS for NOAA operational forecasts, DUNE AI global inferences calculated just for the US, and DUNE AI inferences trained just for the US (b) Heatmap of the HSS for NOAA operational forecasts and DUNE AI inferences.}
  \label{fig: HSS_Comparison}
\end{figure}

Figure \ref{fig: HSS_Comparison} presents a comparative analysis of the Heidke Skill Score (HSS) between NOAA operational forecasts and DUNE AI inferences over 5 years (2019-2023). This comparison highlights performance differences between the two forecasting approaches. The blue curve represents inferences from global DUNE AI results tailored to the United States. In contrast, the green curve represents the DUNE AI model trained by setting all global values over oceans and land, excluding the US, to zero. This is intended to focus training exclusively on the US. Over the five years, the DUNE AI model demonstrates superior performance in February, May, and October but shows reduced skill in March, June, November, and December. The results for January, April, July, August, and September are comparable between the two methods. It is worth noting that NOAA operational forecasts are provided at a 2\textdegree x 2\textdegree resolution, while DUNE AI inferences are available at a finer 0.25\textdegree x 0.25\textdegree resolution.

\begin{figure}[H]
  \centering
  \includegraphics[width=0.8\textwidth]{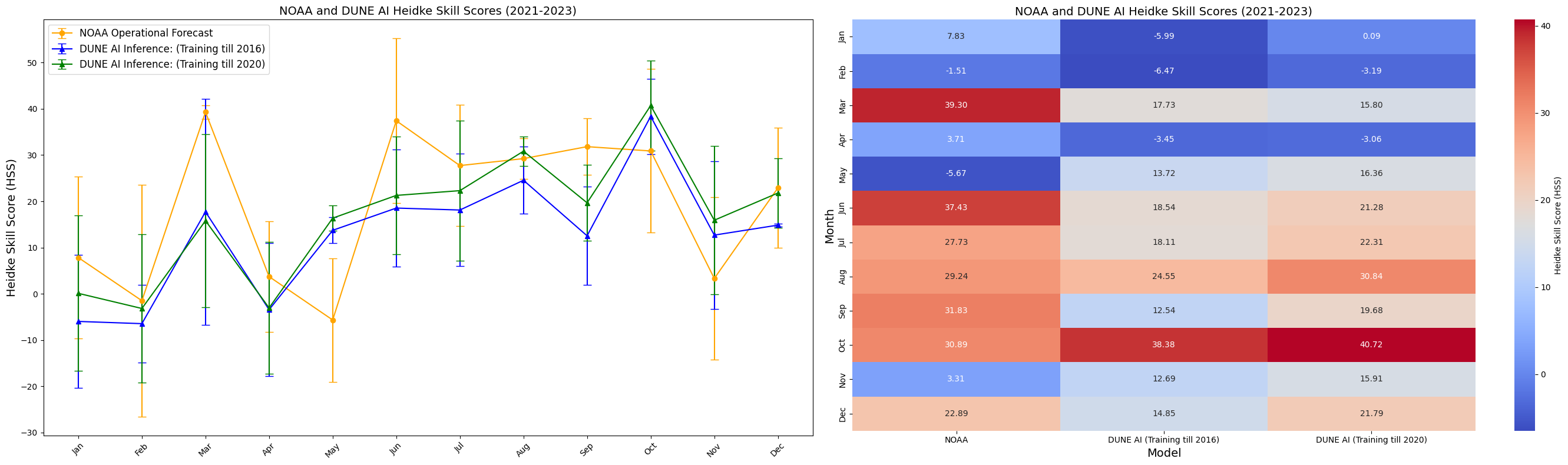}
  \caption{Heidke Skill Score (HSS) Comparison for Monthly Averaged Forecasts (2021-2023) (a) Comparison of monthly averaged HSS for NOAA operational forecasts, DUNE AI inferences trained on data up to 2016, and DUNE AI inferences trained on data up to 2020. (b) Heatmap of the HSS for NOAA operational forecasts and DUNE AI inferences.}
  \label{fig: HSS_Latest}
\end{figure}

Since the DUNE global model was trained on ERA5 from 1980 to 2016, an additional experiment was performed by training DUNE from 1980 to 2020. Global inferences were carried out for 2021 and 2023 without any validation. Figure \ref{fig: HSS_Latest} compares the monthly Heidke Skill Score (HSS) for each month of the 3 years 2021, 2022, and 2023 between NOAA CPC operational Gridded Monthly Forecasts and DUNE AI inferences. The blue curve represents inferences derived from global DUNE AI results tailored to the United States, with the model trained on data up to 2016. The green curve indicates inferences derived from global DUNE AI results tailored to the United States, with the model trained on data up to 2020. The results suggest that training the DUNE AI model with more recent data improves the HSS. This may be a result of continued increased global monthly warming from the period 2017-2020. Based on these results, the DUNE AI model demonstrates comparable performance to NOAA operational forecasts in February, April, June, August, and December. The DUNE model exhibits improved performance in May, October, and November. However, it shows a lack of skill in March, June, and September.

\subsection{Ensemble Members Comparison}
\label{sec: Ensemble Members Comparison}

The ECMWF ERA5 monthly dataset comprises an ensemble of ten averaged monthly members, each with a spatial resolution of 0.5\textdegree x 0.5\textdegree for all available years. To achieve a finer resolution of 0.25\textdegree x 0.25\textdegree, bilinear interpolation was employed to upscale the data. The DUNE AI model was trained on the original monthly-averaged t2m reanalysis dataset. After training, the model's inferred weights were applied to the respective extrapolated dataset for each ensemble member.

\begin{figure}[H]
  \centering
  \includegraphics[width=1\textwidth]{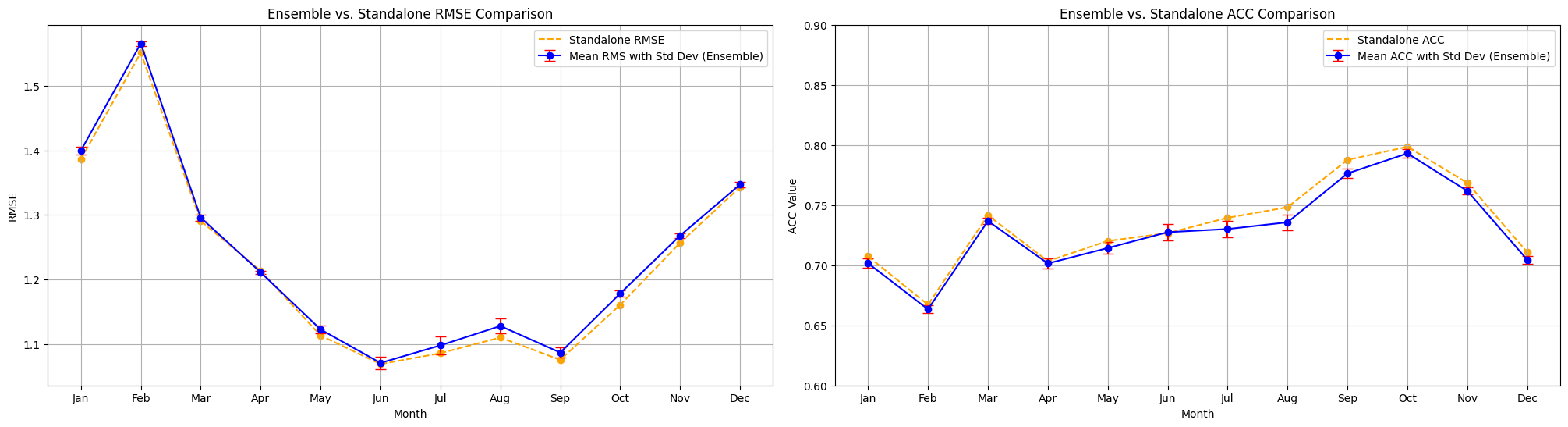}
  \caption{(a) Comparison of RMSE: 10-Member Ensemble vs. Standalone Inference for Averaged Inferences (2019-2023). (b) Comparison of ACC: 10-Member Ensemble vs. Standalone Inference for Averaged Inferences (2019-2023)}
  \label{fig: Ensemble_Comparison}
\end{figure}

The testing period spanned from 2019 to 2023, covering 60 months. Given that there are 10 ensemble members per month and data for 5 years, the total number of sample months for training amounts to 600. The RMSE and ACC were computed for each ensemble member relative to the ERA5 monthly observations and plotted, showing the mean and standard deviation of the ensemble alongside the monthly mean of the original DUNE-inferred RMSE and ACC. Figure \ref{fig: Ensemble_Comparison} illustrates that the variance in RMSE and ACC of the ensemble forecast is only marginally different from the original stand-alone forecasts, with some noticeable deviations exceeding one standard deviation from July to September for ACC and from August to October for RMSE. The conclusion drawn from this analysis is that interpolating ensemble member data from a lower to a higher spatial resolution does not significantly differ from the mean sample of five years. This suggests that extrapolating T2m from the ERA5 resolution to 0.01 degrees may not substantially differ from observations at that higher resolution.

\section{Discussions}
\label{sec: Discussions}
Examining ERA5 data from 1940, observations indicate there is less global warming in temperature from 1940-1979 than appears in the data from 1990 to 2023. Based on these trends, it was not clear which portion of the dataset should be chosen as climatology for use in forming the anomalies in the training of DUNE. As a result, we conducted experiments, not shown herein, that trained the DUNE model on data from the following periods: 1940-2016, 1950-2016, 1960-2016, 1970-2016, and 1980-2016. The results showed that training from 1980-2016 yielded the most accurate global statistical results using the climatology (1950-1979).

Combining SST data over oceans with T2M data over land produced improved DUNE forecast performance, particularly in coastal regions. This improvement can be attributed to SST data providing more accurate and relevant information over oceans, capturing oceanic influences and variations more effectively. Conversely, T2M data is more precise over land areas, reflecting terrestrial temperature changes more accurately. By leveraging the strengths of both datasets in their respective domains, the model better accounts for the complex interactions between oceanic and terrestrial climates, leading to enhanced prediction accuracy. As a result, SST over oceans was adopted in all subsequent experiments reported.

Incorporating constant fields into the model training significantly enhanced the model's overall forecasting capability. These constant fields provide essential static information that aids the model in understanding the underlying physical geography and land surface characteristics, which are crucial for accurate climate predictions. To understand their individual effects, each constant was run as a separate experiment with a separate channel. This approach allowed us to isolate and evaluate the contribution of each field. It was concluded that using the six constant fields helped DUNE better generalize and improve its forecasting results by providing these static environmental factors.

Experiments using the sliding window data from the prior 12, 6, 4, 3, 2, and 1 months to predict the next month's mean temperature did not significantly deteriorate model performance. This method allows the model to capture temporal patterns and trends over longer periods, which is beneficial for multi-month forecasting. As a result, it was included in the presentation here of the DUNE model. This served as an addition of an aspect of attention that is planned for the next version of DUNE. Based on this, it was decided to use only the prior month to forecast the next month, as it provided the most pertinent and actionable data for our predictions. However, for the moving window approach, we used data from the prior x months to forecast the temperatures for the next x months.

Incorporating the average of the four internal ensembles into the DUNE AI model significantly improved forecast accuracy, particularly during winter months when temperature variability is the highest over land. The availability of ten ensemble forecasts allowed for testing the impact of external ensemble forecasting in minutes, which we demonstrated helps address the impact of model uncertainty errors and captures a broader range of possible outcomes. Adding this capability in the future will enhance the robustness and reliability of our forecasts.

In analyzing the monthly mean results published by \citet{taylor2022deep}, it was found that the DUNE model performed equally well, if not better, in predicting SST over the oceans. This comparison highlights the robustness and accuracy of the model in handling oceanic temperature predictions. It's important to note that while \citet{taylor2022deep}'s research focused solely on forecasting SST, the results presented here include both land and oceans, offering a more comprehensive approach. This dual focus allows our model to capture interactions between terrestrial and marine climates, further enhancing its predictive capabilities.

In comparing the forecast results of Dune with those of the NOAA Climate Prediction Center Gridded Monthly Temperature Outlook Product, the extensive methodology involved in its operational production should be noted. For example, NOAA monthly forecasts are based on a number of tools, including ensembles of dynamical models, statistical regression forecasts, pattern correlation anomalies, extensive supplementary data sets, and hybrid models involving many organizational outputs. Based on the favorable comparisons of the simple DUNE regional monthly forecast results over the US shown in section 6.6 at considerably reduced time and resources with those of the operational products, it would indicate a strong potential need to include the DUNE AI product or an equivalent AI method among the complement of other components going into the NOAA Gridded Monthly Temperature Outlook.

An interesting serendipitous ensemble forecast result was the conduct of an experiment with an ensemble of 10 DUNE forecasts from 10 ERA5 monthly reanalyses at ERA5 reanalysis 0.5 degrees extrapolated to 0.25 degrees that exhibited comparable forecast accuracy with forecasts initialized from the ERA5 0.25 degrees. This might indicate the ability of DUNE AI to be independent of the need for resolution retraining when applied to extrapolated higher-resolution data sets. 

In a note of caution, it should be pointed that a limitation inherent to AI/ML climate forecasting relative to physics-based forecasting is its dependence on the training data set selected. DUNE employs atmospheric states of monthly anomalies, which depend on the choice of a climatology period. However, monthly atmospheric states are not in equilibrium, as Figure 13 shows that all months are experiencing global warming. This is particularly true when applied to regional predictions for different domains whose inputs were trained as anomalies based on a specific set of years for defining a climate. The experiment shown in Figure 12 reflects the improved HSS accuracy that can be obtained by selecting more recent climatology for training. In addition, the impact on regional forecast performance can be improved substantially in terms of HSS over the US when trained with data just over the US rather than globally. Future regional forecasts will need to be separately trained.

\begin{figure}[H]
  \centering
  \includegraphics[width=1\textwidth]{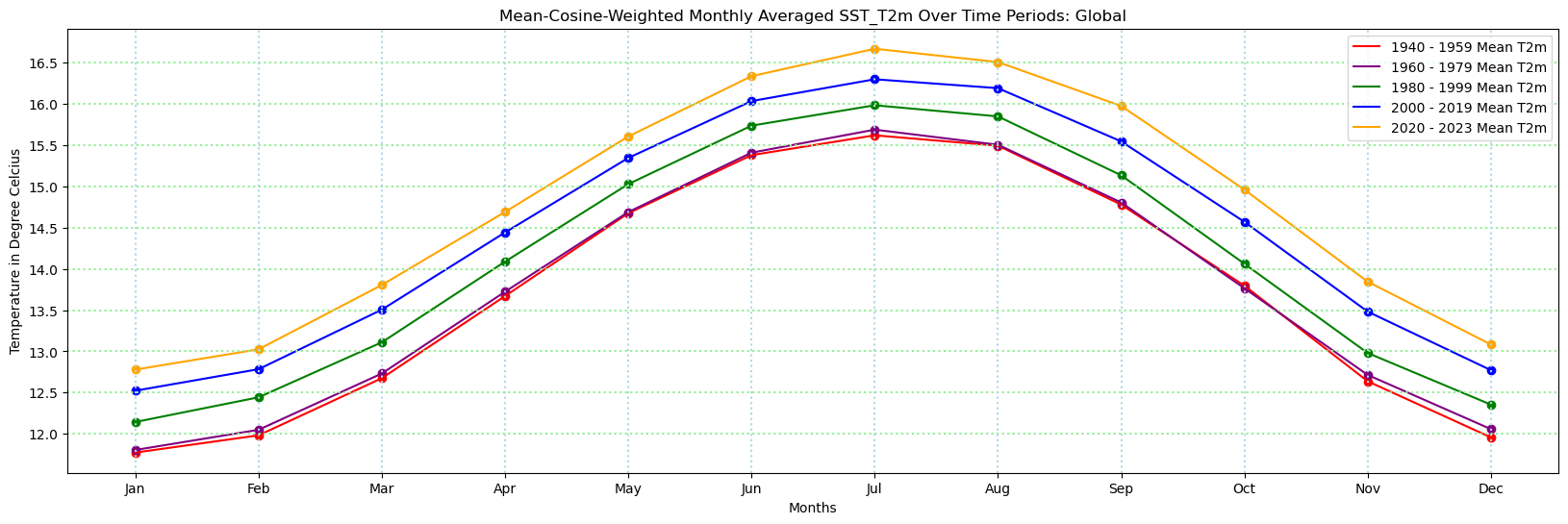}
  \caption{Mean-Cosine-Weighted Monthly averaged global temperatures derived from ERA5. It shows the increase in global temperatures over the years. }
  \label{fig:monthly_averaged_temperature}
\end{figure}

\section{Conclusion}
\label{sec: Conclusion}
A deep learning UNet++-based self-forming ensemble AI model, DUNE, was developed to reduce forecast uncertainty during periods of atmospheric variability in monthly or seasonal near-surface air temperature (T2m) forecasts over global land-based regional domains. The efficacy of DUNE was assessed by comparing its monthly and seasonal RMSE, ACC, and HSS statistics with those of persistence, climatology, and other published physics-based and AI forecast models over multiple years. In all cases, DUNE's error statistics were comparable to or outperformed those of other forecast methods. Additionally, this study presents favorable comparative RMSE, ACC, and HSS error statistics for monthly gridded 2-meter temperature forecasts over the US and other regional domains for the most recent five years, at a resolution of 0.25 degrees compared with the NOAA operational gridded monthly Outlook at 2-degree resolution. The results show that the DUNE AI model's performance is comparable to NOAA's forecasts, as evaluated against ERA5 monthly observations. In addition, for the first time, AI-based Seasonal and Annual forecasts are presented for 2024 pending verification.

\section*{Data Availability Statement}
\label{sec: Data Availability Statement}
The ERA5 monthly mean data used in our study is available for download from Copernicus Climate Change Service \url{https://cds.climate.copernicus.eu/cdsapp#!/dataset/reanalysis-era5-pressure-levels-monthly-means?tab=overview}. This dataset includes all the variables required for our DUNE AI model. The NOAA monthly forecasts are available on their website at \url{https://www.cpc.ncep.noaa.gov/products/predictions/long_range/tools/briefing/mon_veri.grid.php}. The code for training the monthly mean, seasonal mean, annual mean, and moving window approaches is available on GitHub: \url{https://github.com/Pratik-Shukla-1/DUNE_AI_Model}.

\section*{Acknowledgments}
We want to express our gratitude to the NASA/ESTO Firesense Program Manager, Haris Riris, and his staff for their support on our grant number 80NSSC22K1405, which has allowed graduate student support to pursue this critical research project. Their belief in the significance of this AI/ML study in advancing knowledge in seasonal and annual prediction will be instrumental in addressing the impact and risk of wildfires on climate change.

We would also like to acknowledge the support from the NASA HPC office in making the GSFC/NCCS computing facility available to this grant, without which these breakthrough findings would not have been possible. Moreover, we would like to thank the NCCS staff for their professional management in providing reliable system support and access to their advanced machine learning system. That support has been vital in facilitating our data aggregation, training, and predictions of the extensive number of experiments needed to test and evaluate this unique DUNE AI/ML Earth system forecasting model.

Finally, we wish to recognize Prof. Karuna Joshi and the support provided by the UMBC NSF-funded Center for Accelerated Real Time Analytics (CARTA) for providing their Computing and Laboratory resources to support this machine learning collaborative research study.

\begin{appendices}
\renewcommand{\thefigure}{A\arabic{figure}}
\setcounter{figure}{0}

\renewcommand{\thetable}{A\arabic{table}}
\setcounter{table}{0}

\section{DUNE AI Architecture}

\begin{figure}[H]
  \centering
  \includegraphics[width=1\textwidth]{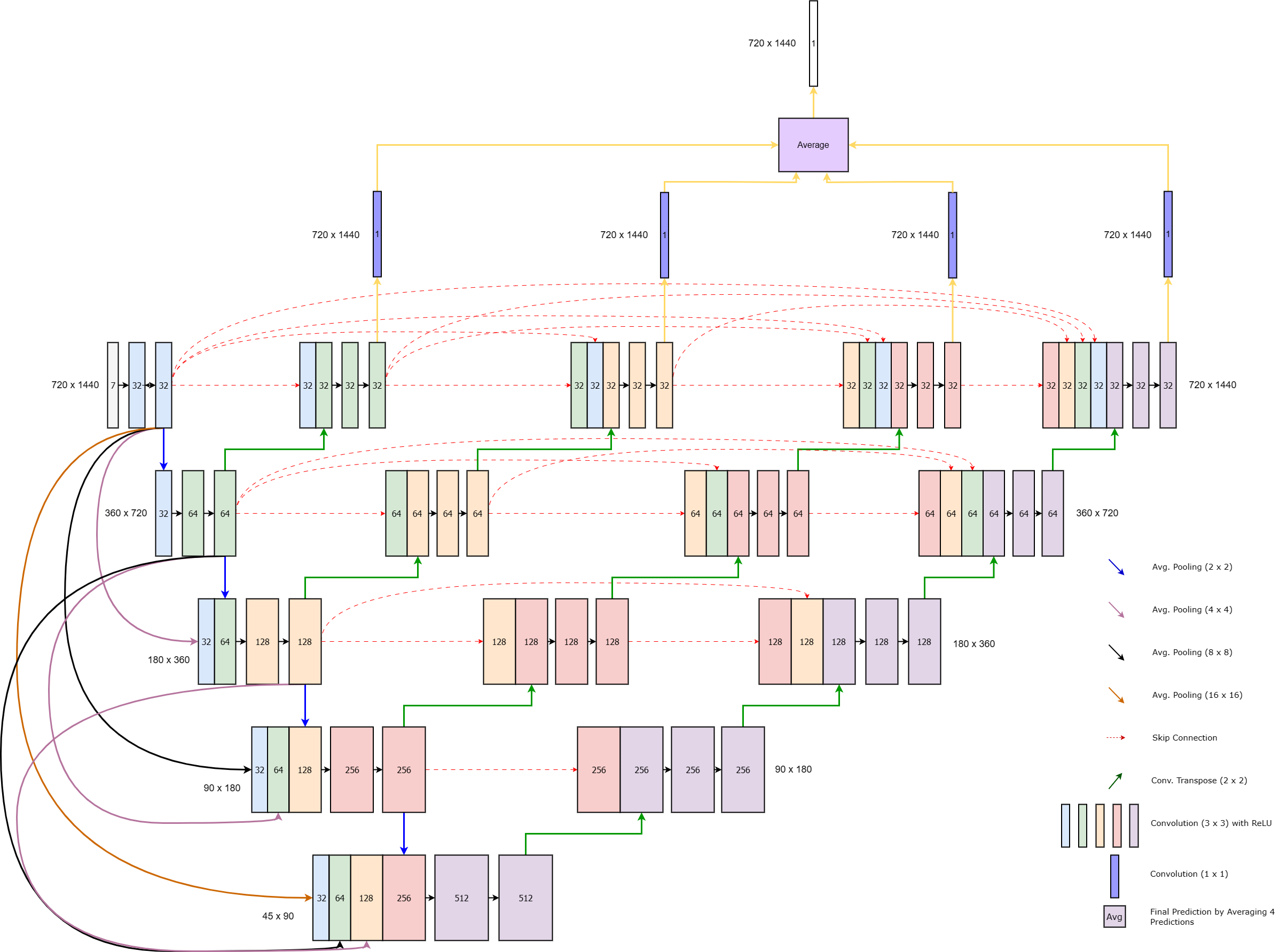}
  \caption{DUNE AI Extended Architecture: The input number of channels is adjusted to forecast the monthly mean temperature one month in advance, depending on the number of time steps (months) to be forecasted. Details are provided in Table \ref{tab:month_vs_channels}. We utilize seven input channels for monthly mean predictions starting one month in advance: temperature, LSM, SLT, Orography, TISR, CVH, and CVL. At the initial level, the input dimensions are 720 x 1440. As the network progresses deeper, the dimensions are halved while the number of channels increases. In the decoding phase of the architecture, the input dimensions are doubled at each level, and the number of channels at each level is indicated within the rectangular boxes.}
  \label{fig: DUNE}
\end{figure}


\section{Extended-Range Forecast Results}
\renewcommand{\thefigure}{B\arabic{figure}}
\setcounter{figure}{0}

\renewcommand{\thetable}{B\arabic{table}}
\setcounter{table}{0}

\begin{figure}[H]
  \centering
  \includegraphics[width=0.5\textwidth]{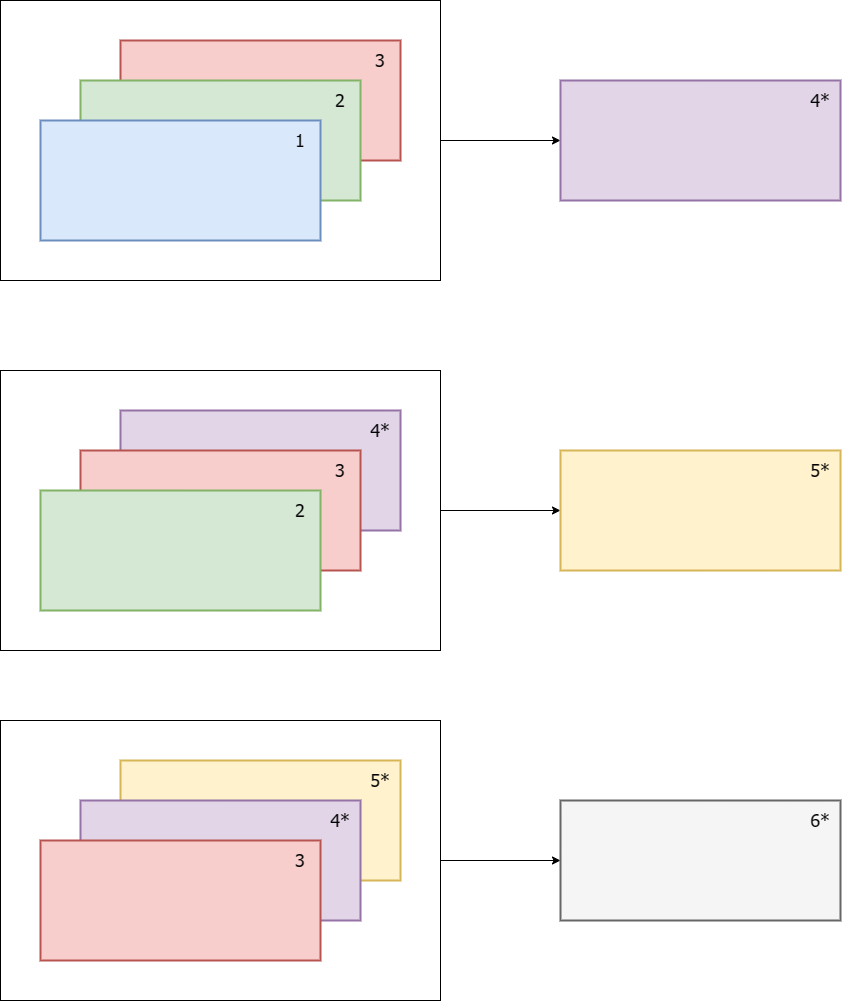}
  \caption{The moving window approach shows the forecast three months in advance. It demonstrates how we can use the ground truth of months 1, 2, and 3 to forecast the temperature for month 4*. Then, it uses months 2, 3, and 4* to forecast the temperature for month 5*. Finally, it uses months 3, 4*, and 5* to forecast the temperature for month 6*. The moving window approach demonstrates how it uses the ground truth of months 1, 2, and 3 to forecast the temperature for months 4*, 5*, and 6*. In this figure, * represents the forecasted month.}
  \label{fig: Moving_Window_Approach}
\end{figure}

Figure \ref{fig: Moving_Window_Approach} illustrates the application of the moving window approach to forecast temperatures three months in advance. This methodology can be similarly extended to forecast temperatures any number of months in advance. The comparative results are presented in section \ref{sec: Monthly Mean Extended-Range Forecast Results}.

Table \ref{tab:month_vs_channels} shows the input number of channels with the respective number of months to be forecasted.  
\begin{table}[hbtp]
\centering
\begin{tabular}{lcl}
\toprule
\textbf{Forecast Window (Months)} & \textbf{Input Channels} & \textbf{Description} \\
\midrule
1 & 7 & 1: temperature, 1: tisr, 5: constants \\
2 & 9 & 2: temperature, 2: tisr, 5: constants \\
3 & 11 & 3: temperature, 3: tisr, 5: constants \\
4 & 13 & 4: temperature, 4: tisr, 5: constants \\
6 & 17 & 6: temperature, 6: tisr, 5: constants \\
12 & 29 & 12: temperature, 12: tisr, 5: constants \\
\bottomrule
\end{tabular}
\caption{Forecast Window (Months) and No. of Input Channels}
\label{tab:month_vs_channels}
\end{table}

\begin{figure}[hbtp]
  \centering
  \includegraphics[width=1\textwidth]{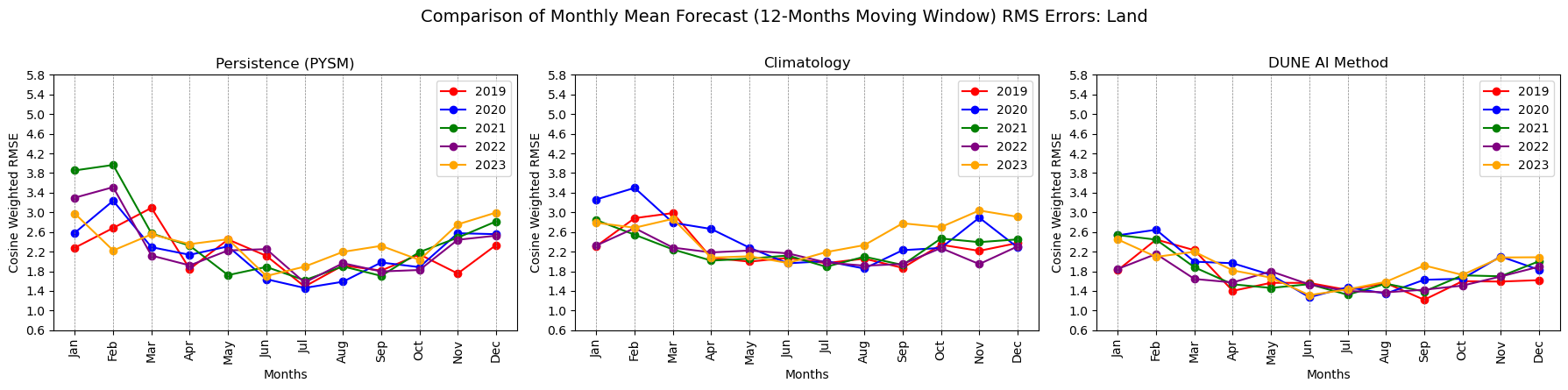}
  \caption{Monthly Mean Forecast (12-Month Moving Window Approach) Land RMSE: (a) Persistence (PYSM), (b) Climatology, and (c) DUNE AI Method. The comparison reveals that the DUNE AI model consistently achieves lower RMSE values, indicating superior forecasting accuracy. Notably, for the record warm year of 2023, the DUNE AI model exhibited relatively small variability in its year-ahead forecasts compared to the baseline models.}
  \label{fig:land_rms_cycle}
\end{figure}

\begin{figure}[hbtp]
  \centering
  \includegraphics[width=1\textwidth]{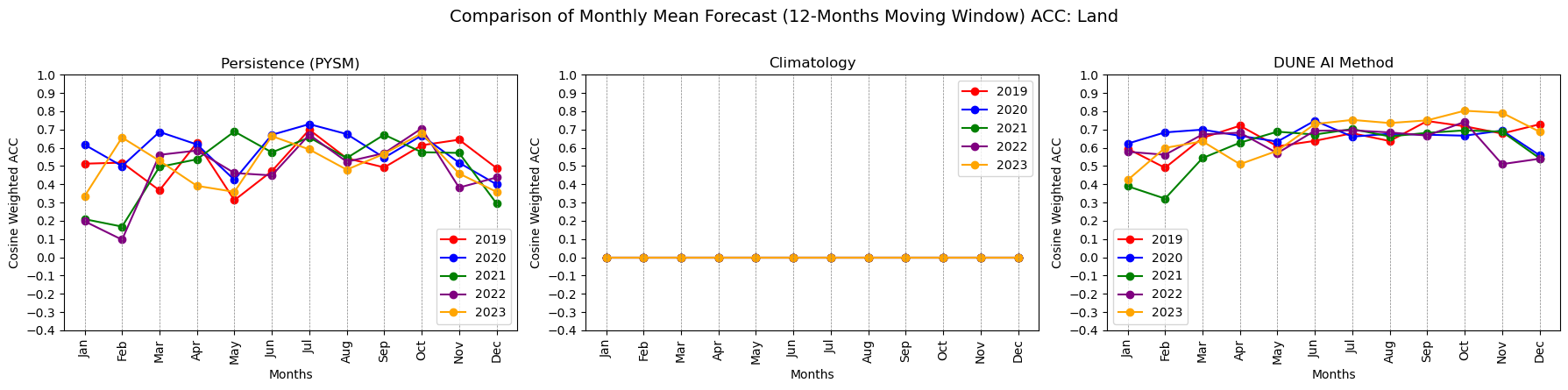}
  \caption{Monthly Mean Forecast (12-Month Moving Window Approach) Land ACC: (a) Persistence (PYSM), (b) Climatology, and (c) DUNE AI Method. The comparison demonstrates that the DUNE AI model consistently achieves higher ACC values, indicating superior forecasting skills. Notably, for the record warm year of 2023, the DUNE AI model exhibited relatively small variability in its forecasts made a year in advance compared to the baseline models.}
  \label{fig:land_acc_cycle}
\end{figure}

Figures \ref{fig:land_rms_cycle} and \ref{fig:land_acc_cycle} display the land RMSE and land ACC for the 12-month moving window over the test period from 2019 to 2023. The DUNE AI model continues to outperform the baseline methods.

\begin{figure}[hbtp]
  \centering
  \includegraphics[width=1\textwidth]{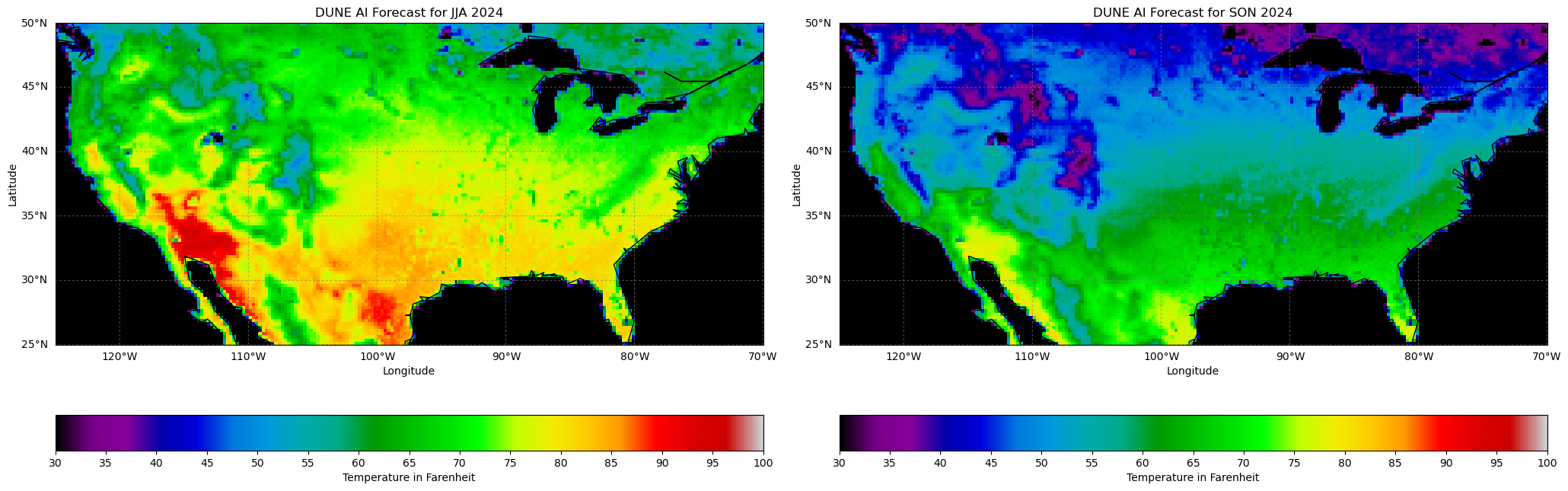}
  \caption{DUNE AI Inference using the 12-month moving window approach for 2024. (a) DUNE AI inference for June-July-August (JJA) 2024 (6 months in advance). (b) DUNE AI inference for September-October-November (SON) 2024 (9 months in advance).}
  \label{fig: US_Moving_Window}
\end{figure}

Figure \ref{fig: US_Moving_Window} shows the forecast of T2m over the US for JJA and SON 2024. These forecast results are pending verification. 

\vfill

\section{Seasonal Mean Forecasts}

\renewcommand{\thefigure}{C\arabic{figure}}
\setcounter{figure}{0}

\begin{figure}[H]
  \centering
  \includegraphics[width=1\textwidth]{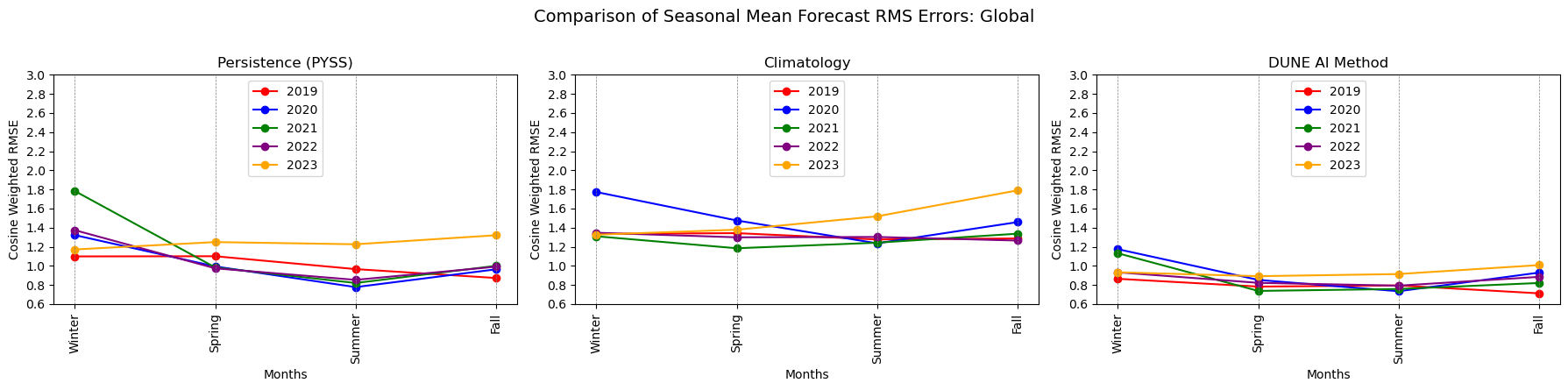}
  \caption{Global RMSE of Seasonal Mean Forecasts: (a) Persistence (PYSS), (b) Climatology, and (c) DUNE AI Method. The comparison shows that the DUNE AI model consistently achieves lower RMSE values, indicating superior forecasting skills. Notably, during the record warm year of 2023, the DUNE AI model exhibited relatively small variability compared to the baseline models. }
  \label{fig:rms_global}
\end{figure}

\begin{figure}[H]
  \centering
  \includegraphics[width=1\textwidth]{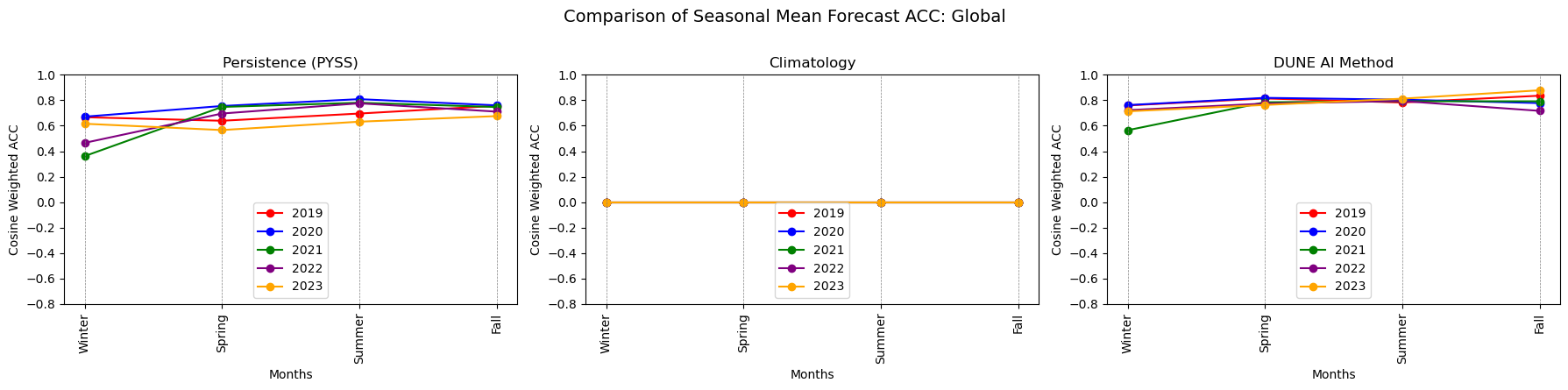}
  \caption{Global ACC of Seasonal Mean Forecasts: (a) Persistence (PYSS), (b) Climatology, and (c) DUNE AI Method. The comparison shows that the DUNE AI model consistently achieves higher ACC values, indicating superior forecasting skills. Notably, during the record warm year of 2023, the DUNE AI model exhibited relatively small variability compared to the baseline models.}
  \label{fig:acc_global}
\end{figure}

\renewcommand{\thetable}{C\arabic{table}}
\setcounter{table}{0}
Figure \ref{fig:rms_global} and Figure \ref{fig:acc_global} show the global RMSE and global ACC for our test period from 2019 to 2023.

\begin{figure}[H]
  \centering
  \includegraphics[width=1\textwidth]{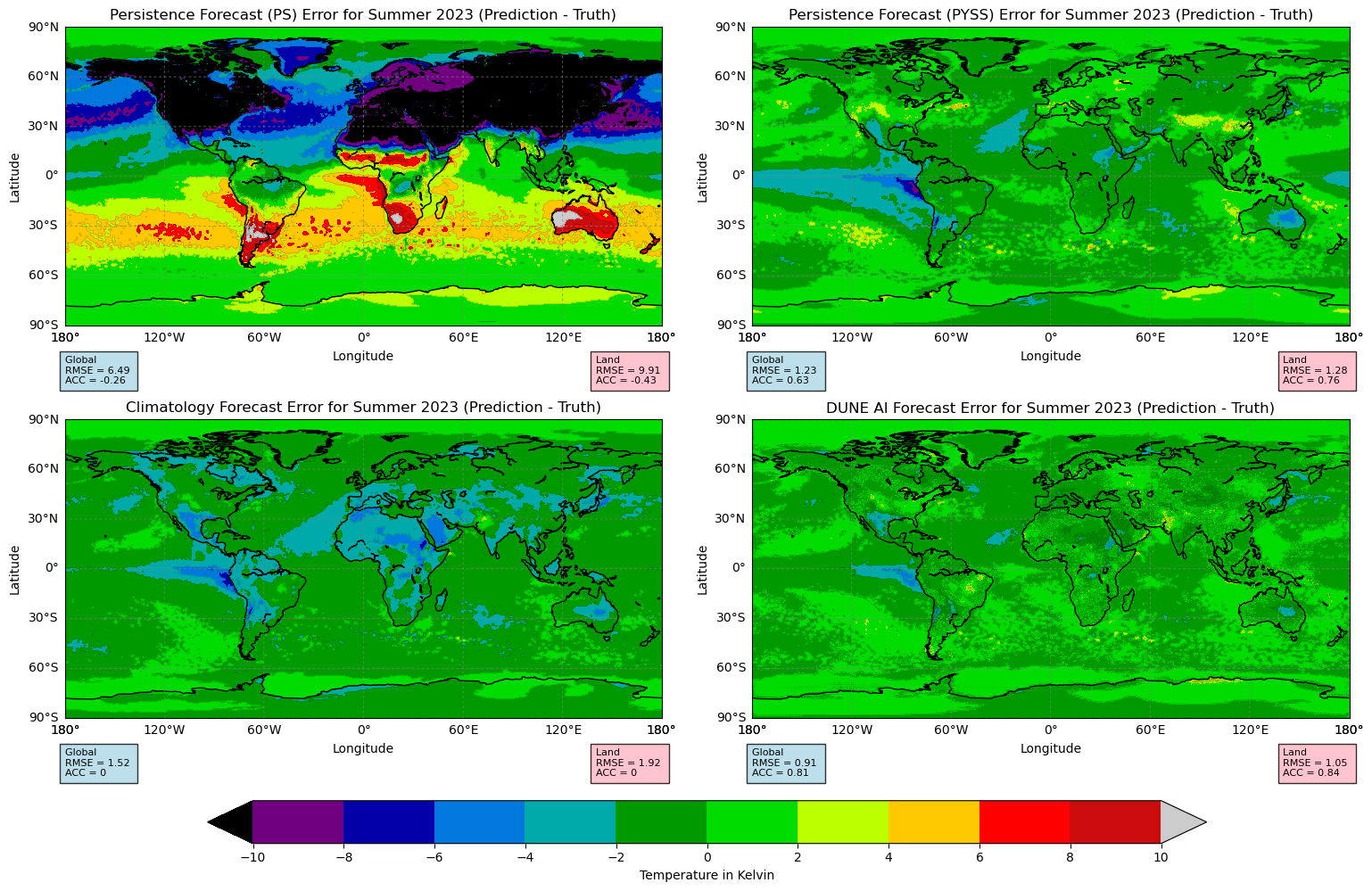}
  \caption{Visualization of forecast errors for summer (JJA) 2023 (a) Persistence (PS) (b) Persistence (PYSS) (c) Climatology (d) DUNE AI Method. The DUNE AI model outperforms other baselines in North America, Africa, Australia, Eurasia, and the El Niño region near South America.}
  \label{fig:global_summer_2023}
\end{figure}

Figure \ref{fig:global_summer_2023} shows the visual comparison of the DUNE AI model forecast errors with the baseline methods for summer 2023.

\section{Yearly Mean Forecasts}

\renewcommand{\thefigure}{D\arabic{figure}}
\setcounter{figure}{0}

\begin{figure}[hbtp]
  \centering
  \includegraphics[width=1\textwidth]{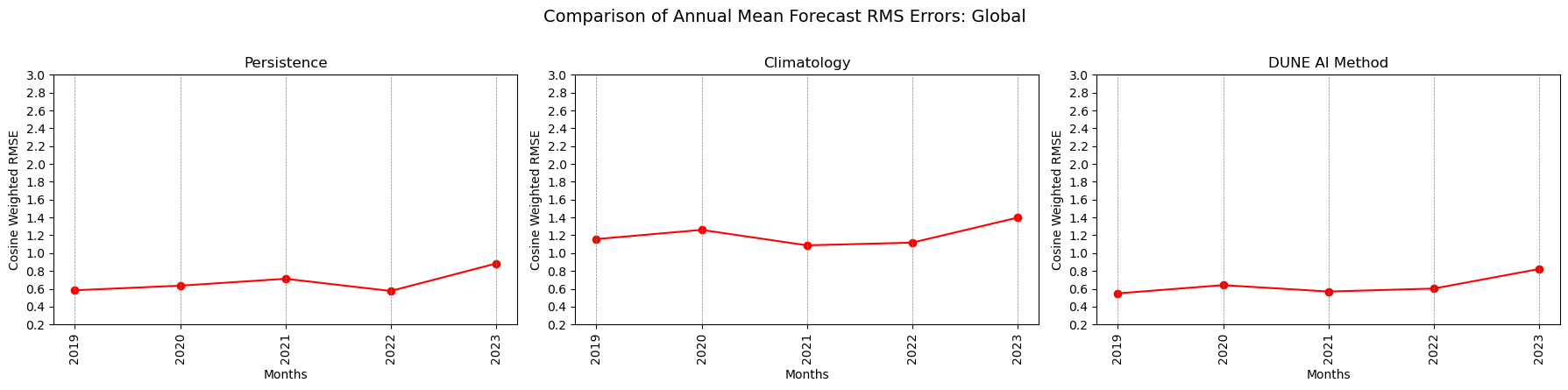}
  \caption{Yearly Mean Forecast Global RMSE (a) Persistence (b) Climatology (c) DUNE AI Method}
  \label{fig:rms_global_yearly_mean}
\end{figure}

\begin{figure}[hbtp]
  \centering
  \includegraphics[width=1\textwidth]{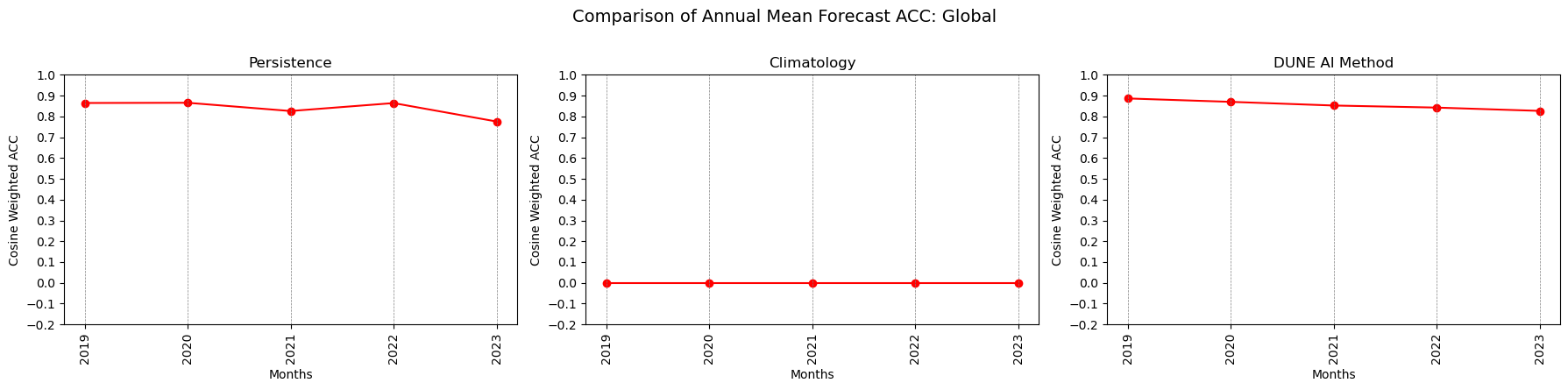}
  \caption{Yearly Mean Forecast Global ACC (a) Persistence (b) Climatology (c) DUNE AI Method}
  \label{fig:acc_global_yearly_mean}
\end{figure}

\renewcommand{\thetable}{D\arabic{table}}
\setcounter{table}{0}
Figure \ref{fig:rms_global_yearly_mean} and Figure \ref{fig:acc_global_yearly_mean} show the global RMSE and global ACC for our test period from 2019 to 2023. 

\begin{figure}[hbtp]
  \centering
  \includegraphics[width=1\textwidth]{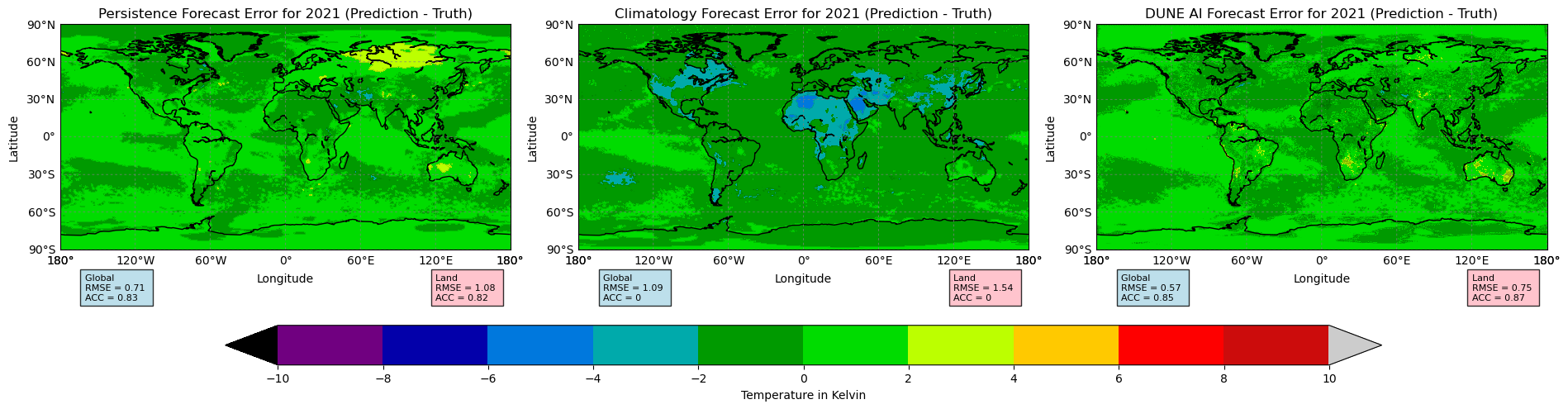}
  \caption{Visualization of forecast errors for 2021: (a) Persistence, (b) Climatology, and (c) DUNE AI Method. The DUNE AI method shows reduced errors in Eurasia, Africa, and North America. Using annual mean data results in better performance compared to monthly mean forecasts averaged over 12 months due to the reduced variability in annual mean predictions.}
  \label{fig:2021}
\end{figure}

Figure \ref{fig:2021} shows the visual comparison of the DUNE AI model forecast errors with the baseline methods for 2021 globally.

\begin{figure}[hbtp]
  \centering
  \includegraphics[width=1\textwidth]{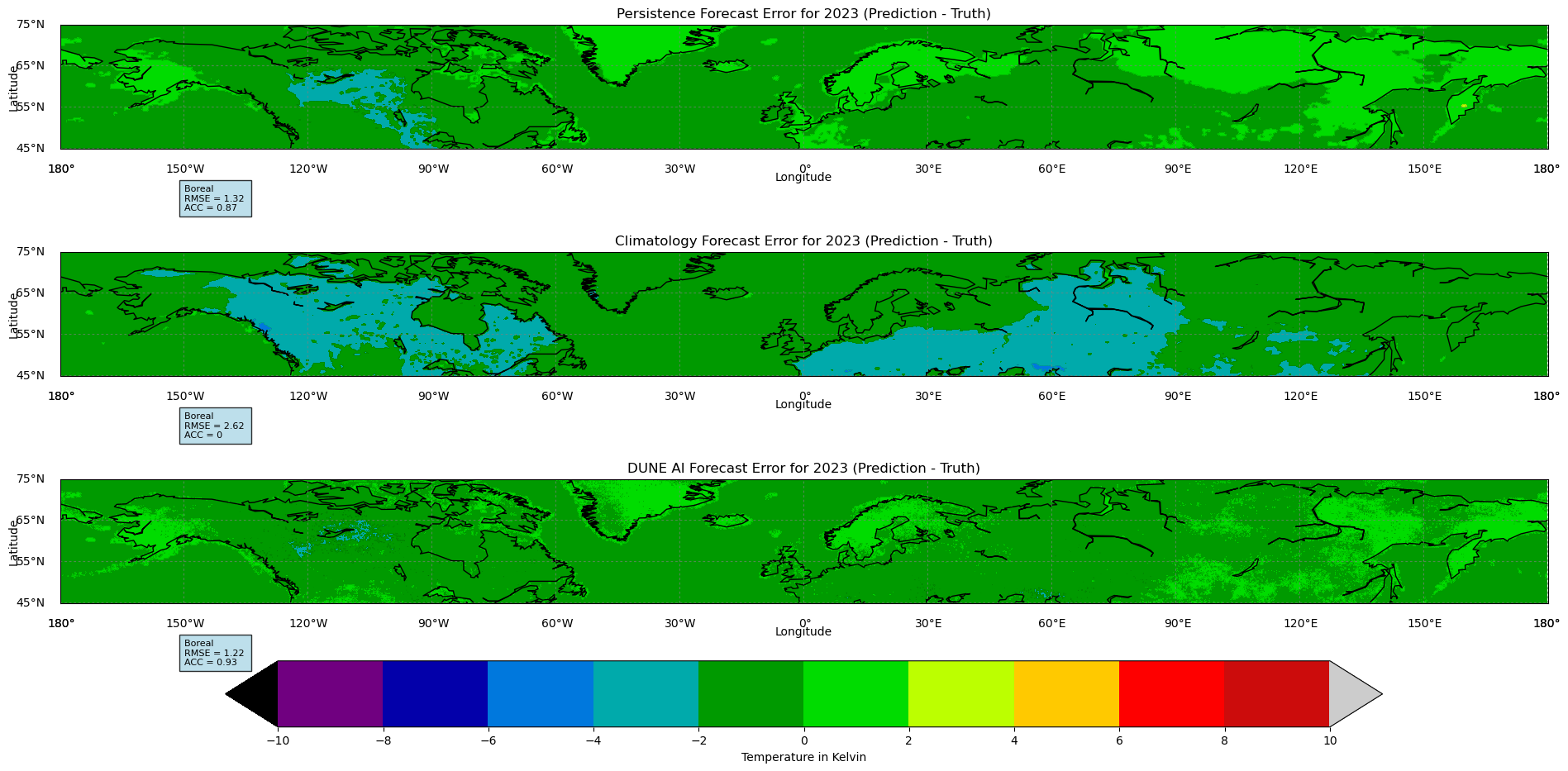}
  \caption{Visualization of annual mean forecast errors for 2023 in the Boreal region (a) Persistence, (b) Climatology, and (c) DUNE AI Method. The DUNE AI method shows the reduced errors in the Boreal region across Eurasia and North America. Africa, and North America. It is worth noting that 2023 was the warmest year on record.}
  \label{fig:boreal_2023}
\end{figure}

Figure \ref{fig:boreal_2023} shows the annual mean forecast errors for the boreal region in 2023. 

\end{appendices}

\end{document}